\documentclass[pdflatex,sn-mathphys]{sn-jnl}
\jyear{2021}
\theoremstyle{thmstyleone}%
\theoremstyle{thmstyletwo}%

\theoremstyle{thmstylethree}%

\usepackage[numbers]{natbib}
\usepackage{amssymb}
\usepackage{arydshln}
\usepackage{lineno} 
\usepackage[mathscr]{eucal} 
\usepackage{amsmath} 
\usepackage{booktabs}
\usepackage{xcolor}
\usepackage{float}
\usepackage{subcaption}
\usepackage{comment}
\usepackage{multirow}
\usepackage[flushleft]{threeparttable}
\usepackage{setspace}

\begin{document}
\let\WriteBookmarks\relax
\def\floatpagepagefraction{1}
\def\textpagefraction{.001}

\title[Auto Response Generation in Online Medical Chat Services]{Auto Response Generation in Online Medical Chat Services}                      


\author*[1]{\fnm{Hadi} \sur{Jahanshahi}} 
\email{hadi.jahanshahi@ryerson.ca}
\author[1]{\fnm{Syed} \sur{Kazmi}} 
\author[1]{\fnm{Mucahit} \sur{Cevik}}


\affil*[1]{\orgdiv{Data Science Lab}, \orgname{Ryerson University}, \orgaddress{\street{44 Gerrard St E}, \city{Toronto}, \postcode{M5B 1G3}, \state{Ontario}, \country{Canada}}}

\abstract{
Telehealth helps to facilitate access to medical professionals by enabling remote medical services for the patients. 
These services have become gradually popular over the years with the advent of necessary technological infrastructure. 
The benefits of telehealth have been even more apparent since the beginning of the COVID-19 crisis, as people have become less inclined to visit doctors in person during the pandemic.
In this paper, we focus on facilitating chat sessions between a doctor and a patient. We note that the quality and efficiency of the chat experience can be critical as the demand for telehealth services increases.
Accordingly, we develop a smart auto-response generation mechanism for medical conversations that helps doctors respond to consultation requests efficiently, particularly during busy sessions. 
We explore over 900,000 anonymous, historical online messages between doctors and patients collected over nine months. 
We implement clustering algorithms to identify the most frequent responses by doctors and manually label the data accordingly. 
We then train machine learning algorithms using this preprocessed data to generate the responses. 
The considered algorithm has two steps: a filtering (i.e., triggering) model to filter out infeasible patient messages and a response generator to suggest the top-3 doctor responses for the ones that successfully pass the triggering phase. 
Among the models utilized, BERT provides an accuracy of 85.41\% for precision@3 and shows robustness to its parameters. }



\maketitle

\section{Introduction}

Online chat services have been used across various sectors for providing customer service, tech support, consultancy/advisory, sales support, and education. 
Compared to in-person and over-the-phone encounters, live chat provides the highest level of customer satisfaction~\citep{econsultancy}. 
As more people join online chat platforms, and with the use of smartphones and smartwatches, as well as an increase in on-the-go communication, smart response generation has become an integral part of online chat platforms.

The smart response suggestions have made businesses more productive as well. Since customer inquiries follow a predictable pattern, which is especially true for domain-specific businesses, smart replies allow for quick and accurate responses. The improved efficiency reduces customers' wait times and thereby results in service satisfaction. Smart response systems also enable employees to handle multiple chats simultaneously, and as a result, businesses can save on additional hiring costs as they grow. 

As healthcare is moving towards online chat services, the smart response system plays a prominent role in allowing smooth and effective doctor-patient interactions. According to the Association of American Medical Colleges (AAMC), the demand for physicians will exceed supply in the U.S. by 2032, leading to an approximate shortage of 46,900 to 121,900 full-time physicians~\citep{aamc}. \citet{mhs} reports that the average wait time for a physician appointment for 15 major metropolitan areas in the U.S. is 24.1 days, representing a 30\% increase over 2014. Furthermore, \citet{patient_visit}'s findings suggest a 60\% decline in the number of visits to ambulatory care practices, whereas there has been a rapid growth in telehealth usage during the COVID-19 pandemic. Due to this high imbalance in the doctor-to-patient ratio and the increase in peoples' reluctance to visit doctors in-person for various reasons (e.g., during the pandemic), telehealth has the potential to become an essential component of our daily lives. 

To facilitate the patient-doctor e-conversations, we develop a smart response generation approach for an online doctor-patient chat service.
We use historical doctor-patient anonymous chats to develop a method applicable to any online doctor consultation service and apps. 
There exist certain challenges regarding these types of datasets. First, in many cases, patients take multiple chat-turns to convey a message, and it needs to be manually determined what part of the chat must be used to match the corresponding doctor's reply. In addition, extensive data preprocessing is required to correct misspellings, punctuation misuses, and grammatical errors.
In our response generation mechanism for the medical chats, we consider various machine learning and deep learning models. Specifically, our algorithm has two steps: a triggering model to filter out infeasible patient messages and a response generator to suggest the top-3 doctor responses for the ones that successfully pass the triggering phase. We observe that response generation mechanisms benefit considerably from the high performance of the deep learning models for the natural language processing tasks at both phases.

The rest of the paper is organized as follows. In Section~\ref{sec:lit_rev}, we present the related literature and summarize our contributions with respect to the previous studies. We define our problem and solution methodology for smart response generation in Section~\ref{sec:smart_response}. Afterwards, we summarize our numerical results with the smart response generator using actual patient-doctor conversations in Section~\ref{sec:results}. The paper concludes with a summary, limitations, and future research suggestions in Section~\ref{sec:conclusion}.

\section{Related work}\label{sec:lit_rev}

The effectiveness of the smart response systems has made them popular in industries where user communication is deemed significant. 
The speed and convenience of simply selecting the most appropriate response make it suitable for high volume and multitask settings, e.g., when an operator has to chat with multiple customers simultaneously. 
A diverse set of suggested options presents users with perspectives they might otherwise have not considered. The correct grammar and vocabulary in machine-generated responses enhance communication clarity and helps users avoid confusion over the context of a message. These attributes can be crucial for businesses that rely on the speed and accuracy of the information and, most importantly, users who lack English proficiency. Additionally, smart reply systems mitigate risks associated with messaging while driving and some health concerns such as De Quervain's tenosynovitis syndrome~\citep{Epstein2020}.     
 
Google's smart reply system for Gmail~\citep{kannan2016smart} serves as a means of convenience for its users. 
With an ever-increasing volume of emails exchanged along with the rise in smartphone use, generating responses on-the-go with a single tap of the screen can be very practical. 
One aspect of the end-user utility discussed in this paper is the diversification of the suggested replies. 
To maximize usability, Google employs rule-based approaches to ensure that the responses contain diverse sentiments, i.e., covering both positive and negative intents. The paper also suggests using a triggering mechanism to detect whether a reply needs to be generated beforehand to save from unnecessary computations and make the model scalable. Uber also devised a one-click-chat model to address driver safety required for responding to customer texts while driving~\citep{weng2019occ}. Their proposed algorithm detects only the intention of the user message, and, using historical conversations, it suggests the most likely replies. The replies are kept short to reduce the time spent reading, and thereby maximizing safety and utility.~\citet{galke2018case} analyzed a similar problem of response suggestion where users of a digital library ask librarians for support regarding their search. They used information retrieval methods such as Term Frequency - Inverse Document Frequency (TF-IDF) and word centroid distance instead of sequence-to-sequence models, noting that such algorithms are more accurate when the training data is limited.

While the models above are task-oriented and designed to accomplish industry-specific goals, they do not address user engagement issues and the motive to make conversations seem more natural. Microsoft XiaoIce used an empathetic computing module designed to understand users' emotions, intents, and opinions on a topic~\citep{li2020}. It can learn the user's preferences and generate interpersonal responses. \citet{yan2018chitty} proposed social chatbot models that serve the purpose of conversing with humans seamlessly and appropriately. \citet{yan2017joint} devised a conversation system in which there were two tasks involved: response ranking and next utterance suggestion. The response ranking aimed to rank responses from a set given a query, whereas the next utterance suggestion was to proactively suggest new contents for a conversation about which users might not originally intend to talk. They used a novel Dual-LSTM Chain Model based on recurrent neural networks, allowing for simultaneous learning of the two tasks. Similarly, \citet{yan2018coupled} designed a coupled context modeling framework for human-computer conversations, where responses are generated after performing relevance ranking using contextual information.

The studies mentioned above discuss the applicability of smart reply and other AI-enabled conversational models in various settings and domains. Smart reply models that are built specifically for online conversations have to adhere to a distinct criterion. In online conversations, users may adopt words and sentence structures differently. The very intention of a user is often expressed and clarified in multiple chats-turns, and responses do not always immediately follow questions or inquiries. To overcome these challenges, \citet{Li2019EnhancingRG} extracted common sub-sequences in the chat data by pairwise dialogue comparisons, which allow the generative model to optimize more on common chat flow in the conversation. They then applied a hierarchical encoder to encode input information where the turn-level RNN encodes the sequential word information while the session-level RNN encodes the sequential interaction information. 

Another challenge with regards to smart reply models built specifically for online conversational chats is scalability. Large-scale deployment of online smart reply models requires energy and resource efficiency. \citet{kim2016} presented the idea of using sentiment analysis to determine the underlying subject of a message, deciding between character vs. word vs. sentence level tokenization, and whether to limit queries to only nouns without affecting the quality of the model. \citet{jain2018evaluating} discussed the idea of using conversational intelligence to reduce both the time and the number of messages exchanged in an online conversation. It includes presenting intelligent suggestions that would engage the user in a meaningful conversation and improve dialogue efficiency. Lastly, \citet{Lee2020} proposed using human factors to enable smooth and accurate selection of the suggested replies.  

Concerning doctor-patient conversation, there have been several studies in recent years to help doctors with artificial intelligence-based diagnostics and treatment recommendations~\citep{hamet2017, topol2019, he2019practical, car2020}. Nevertheless, to the best of our knowledge, there is no specific example of a smart response mechanism in the healthcare domain. Related studies focused on language models such as chatbots and not on real-time chat conversations. \citet{oh2017chatbot} proposed a chatbot for psychiatric counseling in mental healthcare service that uses emotional intelligence techniques to understand user emotions by incorporating conversational, voice, and video/facial expression data. In another study, \citet{kowatsch2017text} analyzed the usage of a text-based healthcare chatbot for the intervention of childhood obesity. Their observations revealed a good attachment bond between the participants and the chatbot. 

As more emphasis is being placed on the quality of patient-physician communication~\cite{Cuffy2020}, an AI-based communication model can facilitate direct and meaningful conversations. However, it is essential to consider the ethical issues related to AI-enabled care~\citep{davenport2019potential, Hancock2020} as well as the acceptability of AI-led chatbot services in the healthcare domain~\citep{nadarzynski2019acceptability}. There is hesitancy to use this technology due to accuracy in responses, cyber-security, privacy, and lack of empathy. 
    
Our study differs from the aforementioned works in the literature in various ways.
For instance, messages exchanged on the Uber platform are typically short and average between 4-5 words. 
The average length of messages exchanged on an online medical consultation service tends to be longer, e.g.,
10 to 11 words on average, and can be up to 100 words,
due to the necessity to clearly describe a certain medical condition.
Similarly, the suggested responses created on Gmail are shorter.
In terms of the corpus size, Uber and Google’s general-purpose datasets might reach millions of instances, whereas a typical training data is substantially smaller (e.g., in tens of thousands), especially for a start-up company or local clinics.
This challenge augments the complexity of our model in that it should learn proper responses using a smaller dataset.
\citet{galke2018case} work with a domain-specific dataset consisting of ~14k question-answer pairs and generate responses using retrieval-based methods. 
On the other hand, their model does not consider diversity and only generates one suggested response for every message. Our model prompts multiple responses with different semantic intents, resulting in better utility for the users. \citet{li2020, yan2018chitty} and~\citet{yan2018coupled} have developed models that are successful in making natural and human-like conversations. 
However, they are generic and not suitable for a domain-specific task such as ours that should take into account the medical jargon.

In this study, we develop an algorithm for smart response generation in online doctor-patient chats. Our analysis is aimed at addressing the challenge of generating smart responses in the medical domain with a limited and constantly evolving dataset (e.g., due to entry of new patients and diseases).
We summarize the contributions of our study as follows.
\begin{itemize}
    \item To the best of our knowledge, this is the first study to propose an auto-response suggestion in a medical chat service. As conversations include medical jargon, we use medical word embeddings and retrain them on our large conversational corpus. 
    
    \item Our method involves employing a novel clustering approach to create a canned response set for the doctors.
    
    \item We employ machine learning algorithms and transformers to address response generation in the medical domain. To the best of our knowledge, it is the first time that BERT, specifically PubMedBERT, is used to generate responses that facilitate chats between patients and doctors. Besides, we use BiLSTM, Seq2Seq, and other machine learning algorithms for performance comparison against BERT. 
    
    \item Our detailed numerical study shows the effectiveness of the proposed methodology on a medical chat dataset. Moreover, our method demonstrates robustness to its parameters. Accordingly, our study provides an empirical analysis of smart auto-response generation mechanisms.
    
    
    \item The proposed method can be used to generate fast smart responses and can be easily integrated into the chat software.
\end{itemize}

\section{Smart Response Suggestion}\label{sec:smart_response} 
AI-assisted tools have become increasingly prevalent in the medical domain over the years. As services such as appointment scheduling and doctor consultations move online, there is an increasing need for auto-reply generation methods to increase the overall system efficiency.
However, as is the case in any domain-specific application, there are certain challenges in developing smart response mechanisms for online medical chat services. 
While particular challenges such as scalability and response quality have been addressed in previous works~\cite{kannan2016smart, weng2019occ}, there has not been much focus on the speed of response generation and disorderly chat flows. We summarize these two issues within the context of a medical chat service as follows.
\begin{itemize}
    \item \textbf{Speed}: As online chats between doctors and patients follow a rapid pace, the model must generate a response instantaneously (i.e., within a second) to be of practical use. 
    This issue does not persist for systems that generate a reply in an offline setting. 
    
    \item \textbf{Disorderly chat flows}: In a chat platform, a message may or may not be followed by an immediate response.
    There are instances where messages are exchanged in various turns with their orders being completely random. 
    A message may be replied to immediately or at a later turn. This issue does not apply to the works that deal with email exchanges, as most of the emails are a direct response to the previous email, or they have a reply-to option to circumvent this challenge. 
    However, the impact of disorderly chat flows might be magnified in doctor-patient chats as the doctors, who might be overwhelmed by conversations, are typically slower than the patients.
\end{itemize}

We address both of these challenges through our comprehensive analysis. 
According to the challenges and the available data, we consider the below-described steps to construct our suggested response mechanism.

\subsection{Data preparation}\label{sec:data_preparation}
In this study, we use a dataset obtained from Your Doctors Online\footnote{\url{https://yourdoctors.online}}, which is an online application that connects patients with doctors. 
The dataset includes a collection of anonymized doctor-patient chats between October 6, 2019, and July 15, 2020. We extracted 38,135 patient-doctor conversations, consisting of 901,939 messages exchanged between them. 
Note that the in-depth data exploratory analysis, e.g., $n$-gram analysis, is excluded due to information sensitivity. 

\begin{figure}[!ht] 
    \centering
    \includegraphics[width = \linewidth]{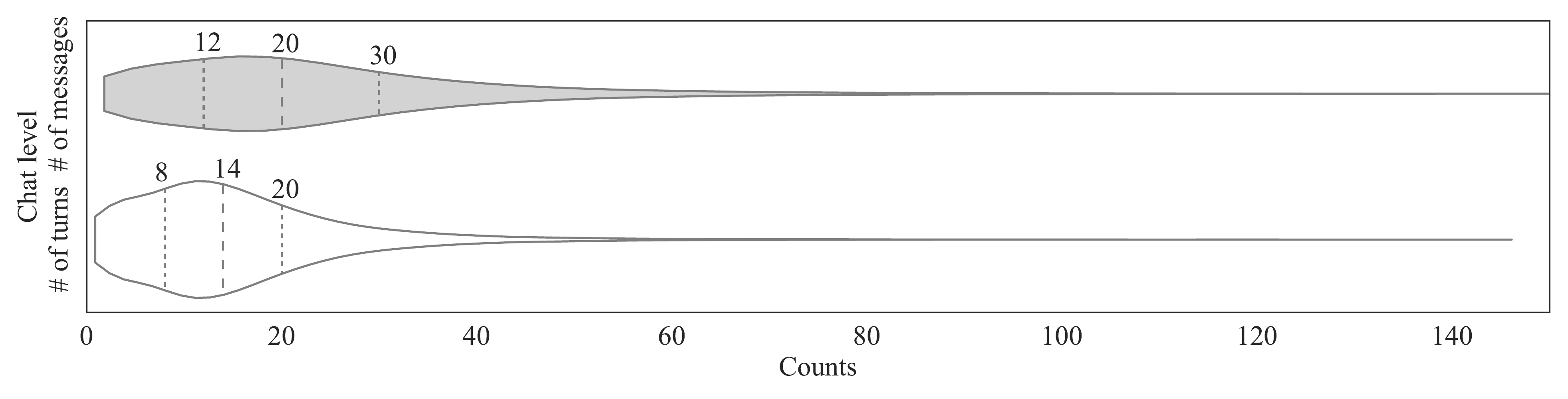}
    \caption{The distribution of the number of turns and messages per chat \textcolor{black}{(Quartile 1 (Q1), median and quartile 3 (Q3) values are shown in the plot.)}}
    \label{fig:n_msg_turns}
\end{figure}

Each chat between a patient and a doctor has two characteristics: the number of messages and the number of turns, i.e., the back and forth messages between them.
The violin plot in Figure~\ref{fig:n_msg_turns} shows the distribution of the number of turns and messages per chat. 
The number of turns in each chat has a distribution with a mean of 15.5 and a standard deviation of 11.5. On the other hand, each chat includes on average 23.8 messages with a standard deviation of 19.3. 

\begin{figure}[!ht] 
    \centering
    \includegraphics[width = \linewidth]{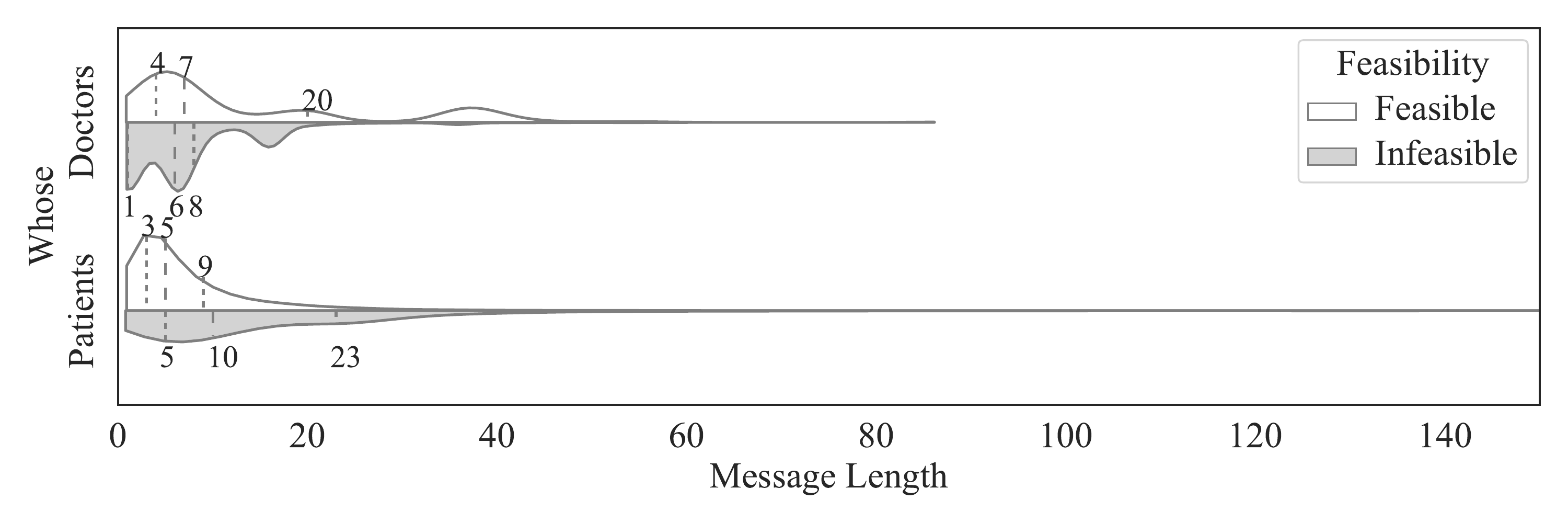}
    \caption{The distribution of the number of words in each message (Quartile 1 (Q1), median and quartile 3 (Q3) values are shown in the plot.)}
    \label{fig:ChatLenDist}
\end{figure}

In the next step, we divide the chat into pairs of patient-doctor messages. Each paired message is manually labeled as ``feasible'' or ``infeasible'', indicating whether a paired message should trigger a smart reply or not. 
Figure~\ref{fig:ChatLenDist} shows the distribution of feasible and infeasible patient-doctor messages' lengths.
We note that the average length of infeasible messages is 20.3 (and $\sigma = 46.3$), whereas the average length of feasible messages is 11.2 (and $\sigma = 7.9$), indicating a significantly higher length for the infeasible ones. 


\subsubsection{Data cleaning}
We define the following filtering conditions to maintain high-quality messages.
\begin{itemize}
    \item We remove any patient/doctor message that is longer than 200 words. That is, we choose not to trigger any responses for those long messages. 
    
    \item As we face a chat environment, there is a plethora of idioms, abbreviations, and mispronounced vocabularies. Therefore, we create a dictionary of abbreviations and replace each abbreviated word with its long-form, e.g., ``by the way'' as a substitute for ``BTW'' or ``do not know'' in place of ``dunno''. Moreover, using the ``pyspellchecker'' package in Python, we generate a comprehensive dictionary of typos in the medical domain. This misspelling dictionary includes 30,295 words extracted from the chats and is used to clean the dataset. Nevertheless, not all the misspelled words are retrievable. We are unable to suggest the proper replacement for some typos that are not similar to any words in the package's corpus.
    
    \item We decide to keep stopwords in the dataset as the final response needs to be grammatically correct. Therefore, we examined our method with and without stopwords and found that keeping them enhances the replies' quality. Similarly, we do not apply lemmatization because it is detrimental to syntactic comprehension.  
    
    \item Other preprocessing steps include removing extra white spaces, deleting punctuation and URLs, converting all characters to lower case, and, finally, removing non-Unicode characters and line-breaks. 
\end{itemize}

\subsubsection{Creating canned response set}
As the response generation task requires labeled data, and considering that pairing patient and doctor messages is a tedious task, we select a portion of the data that captures the most significant characteristic of the desired output. Hence, we divide the work into two folds. First, we explore the similarity between doctors' messages and cluster them, and second, we find the patient pair for each doctor message in only dense, frequent clusters. 

\begin{figure}[!ht] 
    \centering
    \includegraphics[width = 0.60\textwidth]{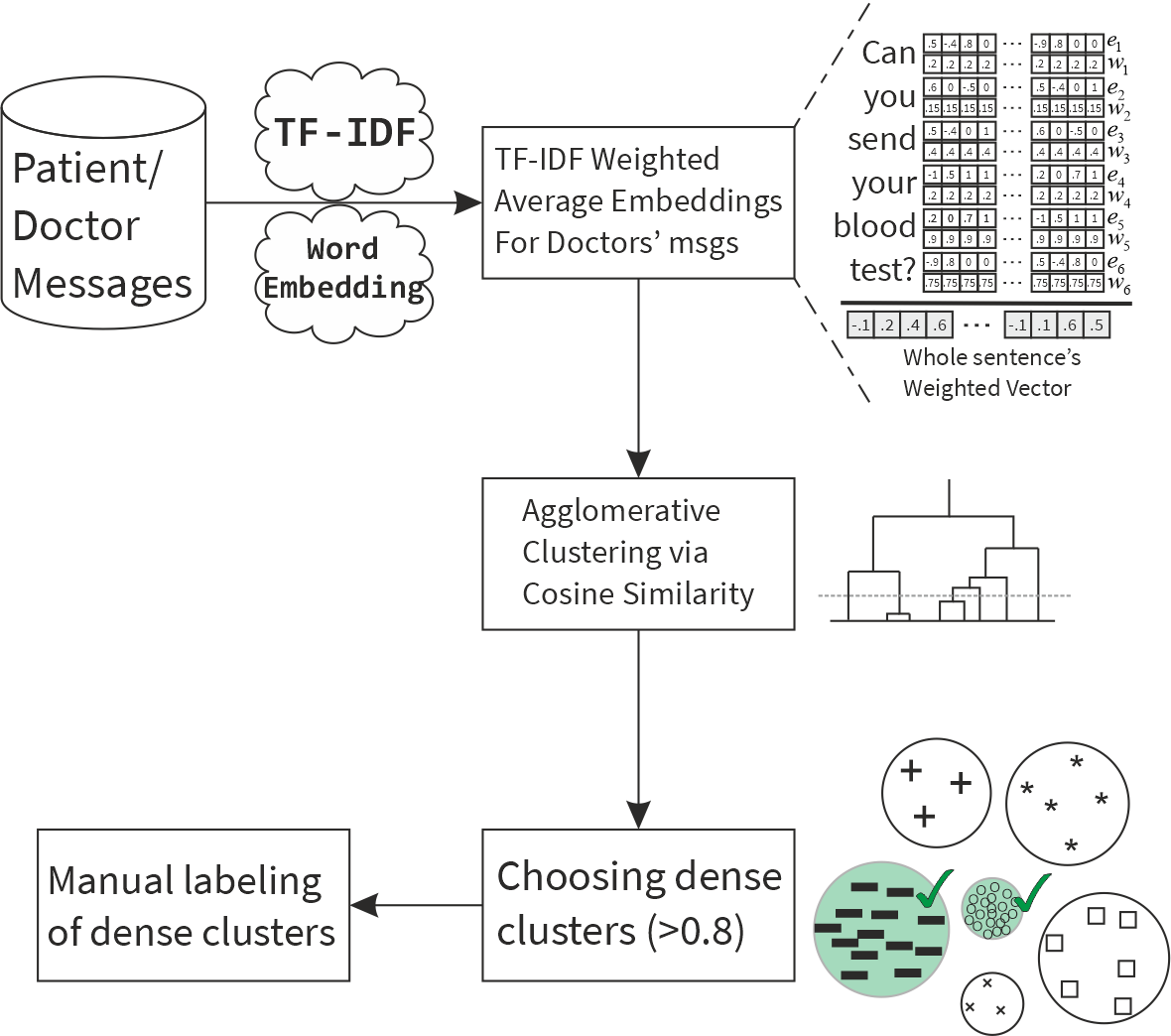}
    \caption{A flowchart for the manual labeling process}
    \label{fig:labeling}
\end{figure}

After data cleaning, we pinpoint the most frequent responses by doctors. However, this cannot be achieved solely by exploring response occurrence since many responses deliver the same message. For instance, ``you're welcome'', ``happy to help'', ``no problem'' and ``my pleasure'' are different possible answers to the same patient message. Therefore, we create a semantic cluster of the responses and examine the total frequency of responses in each cluster. In other words, the model should only learn the messages most commonly sent to the patients. Figure~\ref{fig:labeling} demonstrates the steps 
in the manual labeling process.

As shown in Figure~\ref{fig:labeling}, we convert each textual message to numeric vectors through the weighted average word embedding. 
\mbox{TF-IDF} value for each word generates its weight, and its word embedding is treated as its value. 
As there are many medical terms in the messages exchanged, we use Wikipedia PubMed word embedding\footnote{\url{https://bio.nlplab.org/}}. 
This word2vec dataset is induced on a combination of PubMed and PMC corpus, together with the texts extracted from an English Wikipedia dump. 
Therefore, it is suitable for both medical terms and daily language. 
When compared to the Glove embedding~\citep{pennington2014glove}, which is another popular word embedding, we found that many vocabularies for our task that does not exist in the Glove embedding is supported by Wikipedia PubMed embedding.
Moreover, 
our preliminary experiments
point to a better performance of the Wikipedia PubMed embedding, e.g., the performance is improved by 3.4\% in \textcolor{black}{in terms of precision@1, which measures the performance based on 
the number of relevant responses among the top $k=1$ generated responses}. 
After finding the proper embedding, we use the weighted average word embedding for doctor's messages and apply agglomerative clustering on the responses through the cosine similarity. \textcolor{black}{The words that do not exist in the Wikipedia PubMed word embedding are disregarded while taking the weighted average.} 
Next, using average silhouette width, we found the optimal number of clusters as 158. 
Among them, we chose the clusters whose densities are more than $80\%$. 
Unlike dense clusters that include distinct message types, sparse ones contain a high volume of irrelevant messages with little or no similarity, and hence, we exclude the non-dense clusters. 

We note that the model aims to choose proper responses from the set of canned messages. The smart response generator is not expected to replace a real doctor or act as a bot; instead, its purpose is to facilitate the chat between a doctor and a patient. Therefore, such a canned response set helps doctors respond to frequently asked questions or phrases promptly. In the response suggestion mechanism, having a set of canned messages is a common approach (e.g., see~\citet{kannan2016smart}). 
\textcolor{black}{
The manual labeling task for the patient messages is divided between multiple annotators.
Because our dataset includes a large number of unlabeled instances, each annotator is allocated a distinct subset of the dataset.
Accordingly, we cannot directly measure the inter-annotator agreement rate between the annotators with our current setting.
We note that the proposed labels for each patient message come from a predefined canned response set, which is created based on the collaborative work of the annotators with the doctors.
In addition, nontrivial instances are thoroughly checked by multiple annotators to minimize the chance of having erroneous or conflicting labeling.
}

\subsubsection{Pairing responses and manual labeling} \label{sec:pairing}
\begin{figure}[!ht]
     \centering
     \begin{subfigure}[b]{0.32\textwidth}
         \centering
         \includegraphics[width=\textwidth]{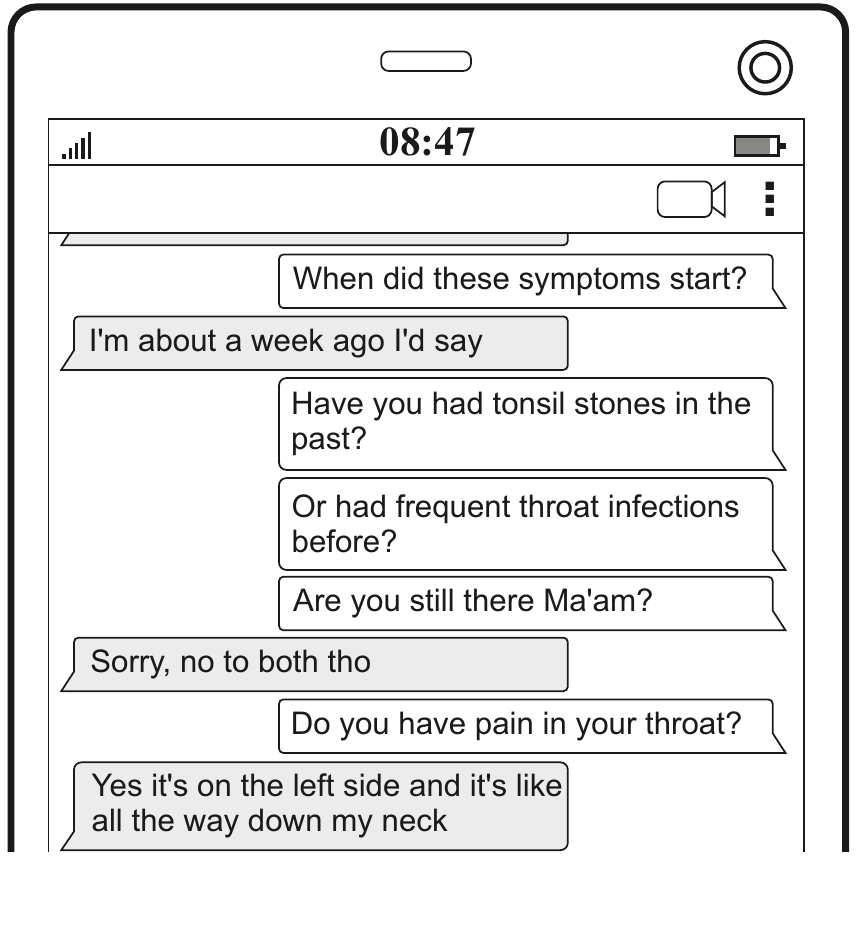}
         \caption{Example 1 - disorderly flow}
         \label{fig:msg1}
     \end{subfigure}
     \hfill
     \begin{subfigure}[b]{0.32\textwidth}
         \centering
         \includegraphics[width=\textwidth]{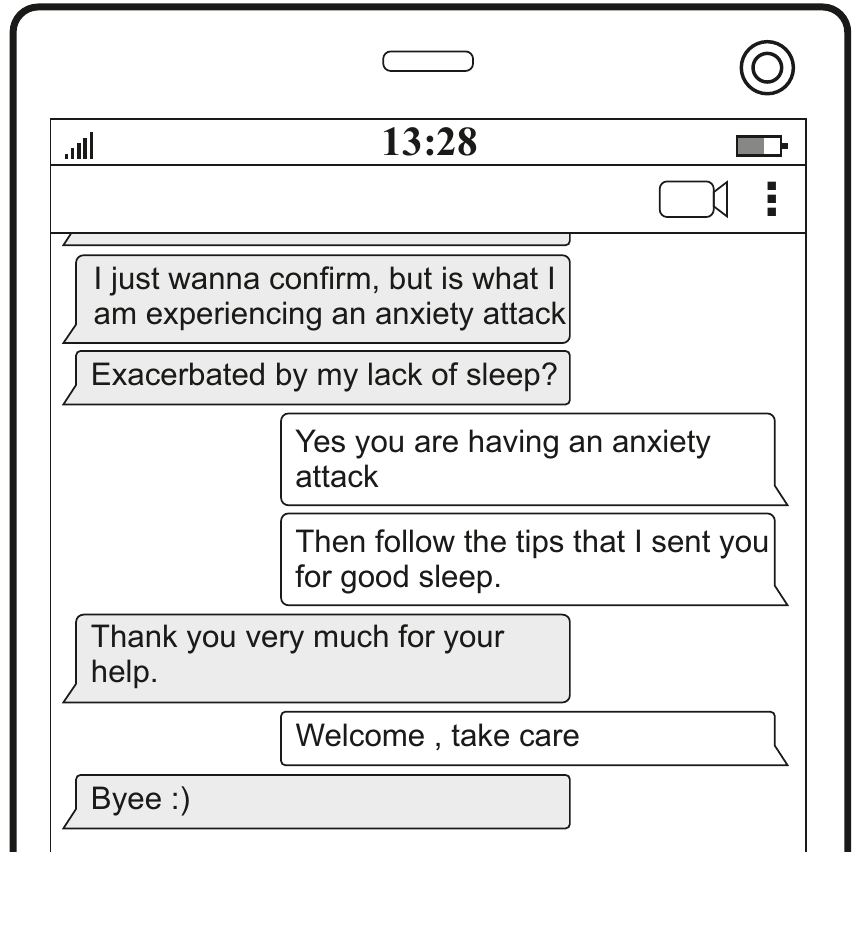}
         \caption{Example 2 - correct flow}
         \label{fig:msg2}
     \end{subfigure}
     \hfill
     \begin{subfigure}[b]{0.32\textwidth}
         \centering
         \includegraphics[width=\textwidth]{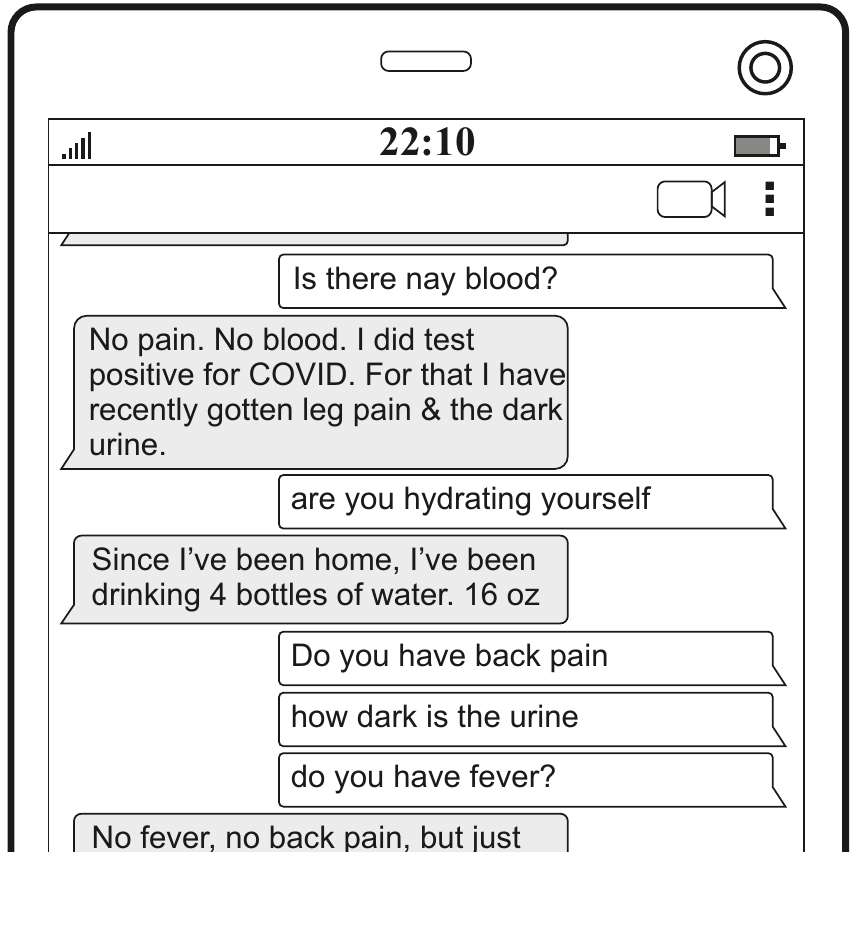}
         \caption{Example 3 - disorderly flow}
         \label{fig:msg3}
     \end{subfigure}
        \caption{Examples of patient-doctor chats; the gray boxes indicate patient messages, and the white ones belong to doctors.}
        \label{fig:examples}
\end{figure}

After obtaining the possible clusters for doctor messages, we pair each doctor message to its related patient message. 
During the manual labeling process, we encounter some challenges in the chat context. 
First, not all messages are a response to their previous message. 
For instance, a doctor may give some information regarding possible drugs without being asked to, or in some cases, a response is too generic or too specific and cannot be considered feasible. 
In such cases, instead of finding the paired patient message, we mark them as ``infeasible''. 
Second, message flow is not always in order. 
For instance, a patient may ask a question, and then in the following messages, give some additional information. 
Then, the doctor starts responding by asking something about the patient's first message. 
As the dataset does not include the ``reply-to'' option, we need to manually trace chats back to find a relevant patient message given each doctor's reply. 
It is a cumbersome task in the labeling process, which does not exist in previous works. 
Figure~\ref{fig:msg1} and Figure~\ref{fig:msg2} show examples of disorderly and correct flow, respectively. 
In some cases, a doctor's response may be relevant to something asked much earlier. 
For instance, in Figure~\ref{fig:msg3}, the question of ``how dark is the urine'' is related to a message sent earlier. 
Also, the doctor's response is related to only parts of the text sent by the patient, and not all of it. 
\textcolor{black}{In such cases, we check the last message by a patient and trim the irrelevant part of the text.
We note that this is done only for the training set, and no trimming is performed for the messages in the test set.}
Afterward, wherever we are unable to find the desired match, we label the message as ``infeasible''. 
Note that the entire manual labeling process is repeated by multiple experts to ensure reaching a consistent canned response set.

Ultimately, the manual labeling process leads to a set of paired patient-doctor chats and some infeasible cases. 
In total, we obtain 31,407 paired messages, 23.1\% of which are ``infeasible''. 
\textcolor{black}{We also employ latent Dirichlet allocation to better understand the characteristics of feasible and infeasible messages by finding the topics within the patients' messages. 
We find that the main topics for the feasible messages include expressing gratitude, informing about diseases, and describing the symptoms. 
On the other hand, topics of the infeasible messages do not follow specific patterns (e.g., providing personal information or thanking the doctor), and they are typically not indicative of requiring a response.}

\subsection{Response diversification}
After finding the appropriate responses and associating them with patient messages, we consider strategies for diversifying the generated responses. 
Based on the comprehensive rules that we identified, we generate a set of diverse canned messages. Note that determining such rules for response diversification requires domain expertise and interaction with the stakeholders (e.g., physicians and end-users). Table~\ref{tab:response_diversification} shows an example of our rule-based response diversification. We diversify the response ``You are welcome'' based on some predefined rules. For instance, if the patient message implies the end of the conversation, we use ``You are welcome. Take care. Bye.'' instead.
As we consider a platform that suggests the top-3 responses to the doctors, our algorithm can benefit considerably from a more diverse set, including all possible situations. 
Otherwise, many irrelevant messages might pop-up on the platform, all pointing to the semantically identical response. 


\setlength{\tabcolsep}{6pt} 
\renewcommand{\arraystretch}{1.25} 
\begin{table}[!ht]
\centering
\caption{An example of rule-based response diversification for the message ``You are welcome.''}\label{tab:response_diversification}
\resizebox{\textwidth}{!}{
\begin{tabular}{ll}
\toprule
\textbf{Diversified response} & \textbf{Adopted rules for the response diversification} \\
\midrule
You are welcome. & A general answer to thanks, thank you, etc. \\
\hline
You are welcome. Take care. Bye. & \begin{tabular}[c]{@{}l@{}}Answer to thanks at the end of conversations.\\ The patient message should imply the end of the chat.\end{tabular} \\
\hline
You are welcome. Have a great day. & When a patient ends the chat by wishing a nice day. \\
\hline
You are welcome. Have a great night. & When a patient ends the chat by wishing a good night. \\
\hline
\begin{tabular}[c]{@{}l@{}} Take care, Happy to help! If you liked \\ our service, please leave us a Google review :) \end{tabular} & When a patient ends the chat implying a satisfactory service. \\
\bottomrule
\end{tabular}
}
\end{table}

\subsection{Smart response generation approach}
In real-time chat conversations, it is typically not required to generate a response for all the received messages, which is unlike a chat-bot.
Therefore, after preprocessing the messages, we define a triggering model that decides whether or not to trigger a reply for a given patient message. 
Triggering is a binary classification task based on the ``feasible''/``infeasible'' manual labeling explained in Section~\ref{sec:pairing}. 
If a patient message passes the triggering model with a prediction probability greater than a predetermined value $p$, then it enters the smart response generator phase; otherwise, we do not generate a reply for it. 

\begin{figure}[!ht]
    \centering
    \includegraphics[width =\textwidth]{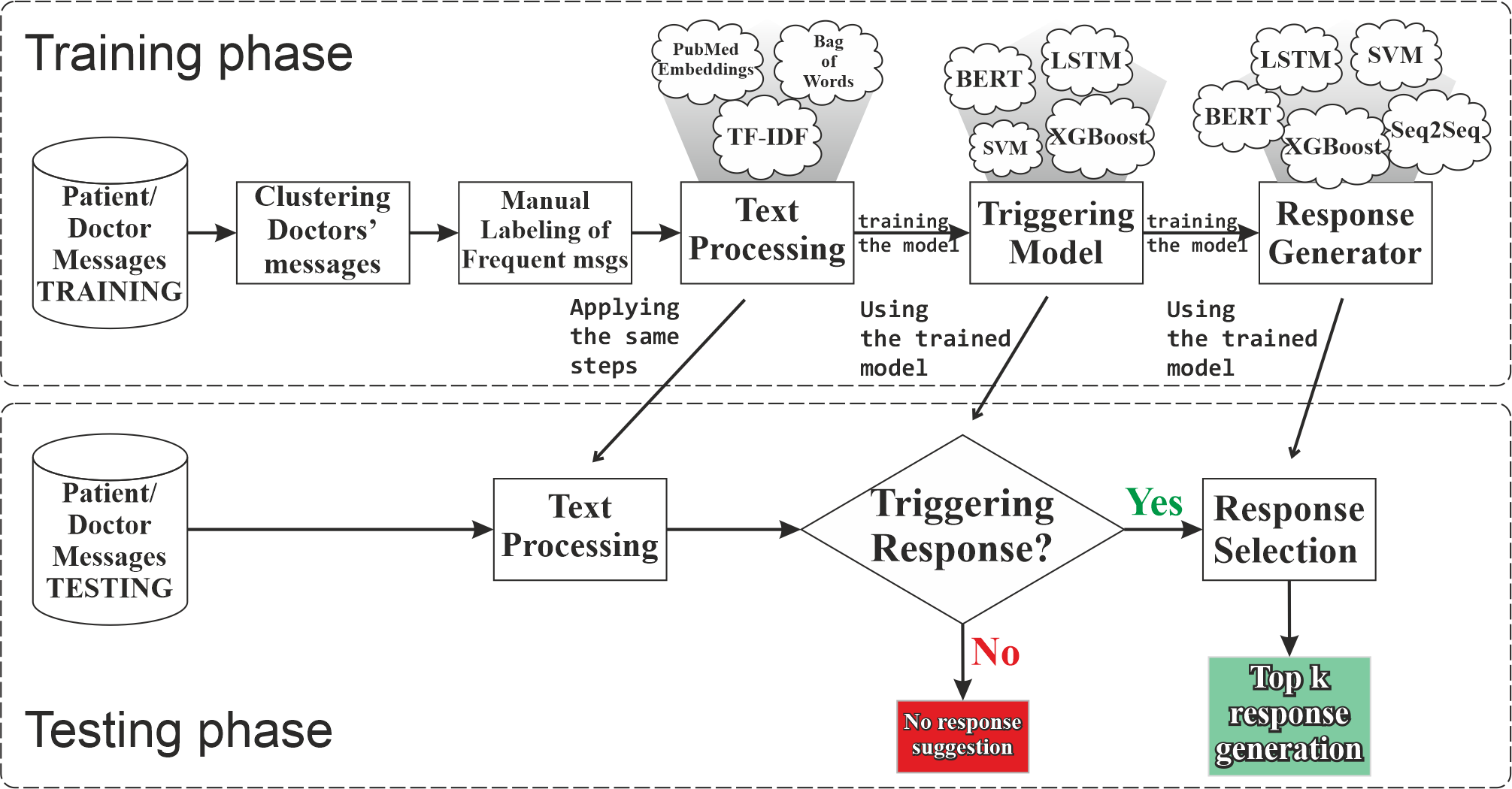}
    \caption{Triggering filter and response generation}
    \label{fig:ExpFlow}
\end{figure}

Figure~\ref{fig:ExpFlow} illustrates the processes of triggering and response generation. 
The reply suggestion phase integrates different models to generate a proper suggested response. 
Since a typical usage in practice involves recommending top-$k$ responses (e.g., $k=3$),
our main aim is to propose the most appropriate response within the first $k$ suggestions. \textcolor{black}{
Note that Figure~\ref{fig:ExpFlow} provides a high-level overview of the entire process, from raw data to generated responses. Specific details regarding the numerical study are provided in Section~\ref{sec:experimntal_setup}. 
As shown in this visualization, triggering and response selection models are trained on the training set, and then the trained models are employed to suggest the proper response for the patient messages in the test set.
While determining the training and test sets, we do not split based on the specific time index of the consecutive messages in the chats to avoid look-ahead bias because we do not treat chat data as time series data. 
That is, each patient message is treated as a separate data instance to create our aggregate dataset, considering that each patient message can be assumed to require an independent response from the others in a chat thread.
Accordingly, stratified random sampling can be employed to create training and test sets.
}

\subsection{Models for triggering and response generation} \label{sec:models}
\textcolor{black}{We examine the performance of different machine learning algorithms and compare their performance with similarity-based and rule-based baselines. 
Below, we summarize the methods employed in our analysis.
}

\subsubsection{Machine learning models} \label{sec:machine_learning}
In the triggering phase, we aim to find the feasibility of the response generation. 
If a patient's message is too specific, (i.e., not applicable to other people), too generic (e.g., ``OK'' or ``done''), or not seen in the training set (i.e., the chance of irrelevant suggestion is high), then there is no need to trigger any smart response. 
Furthermore, the triggering model ignores messages that are too complex or lengthy. 
On the other hand, the system should facilitate a doctor's job since they might be busy with multiple chats. 
The triggering should pass a message to response generation only if a proper response suggestion is likely. 
We experiment with different binary classification methods to identify the most suitable model for the triggering phase.
Accordingly, the value 0 for the dependent binary variable represents patient messages for which it is not ideal to generate a reply, and the value 1 indicates feasible patient messages. 
We use the preprocessed patient messages along with their length as the independent variables. 
The textual feature is converted to numeric values in different ways for each algorithm; therefore, we discuss the data conversion process within each model's explanation.

Unlike the triggering phase that deals with a binary output, the response generation decides on the proper reply only if it passes the first phase. 
\textcolor{black}{
As the set of potential responses is more diverse compared to the binary classification task in the triggering phase, corresponding (multi-class) classification models tend to have lower accuracy in the response generation phase. 
Specifically, we consider the doctor responses as the dependent variable (i.e., target) and the patient messages as the feature values (i.e., predictor).
Hence, a classification model trained for response generation aims to predict the most suitable response for any given patient message.
}
We did not find any significant correlation between the length of a message and the generated response; therefore, we did not include patients' message length as a feature. 

Although there are many machine learning (ML) algorithms for text classification, we chose to experiment with those commonly used in different domains. 
Moreover, in our preliminary analysis, we experimented with other ML methods (e.g., Random Forest and Naive Bayes); however, we did not find those to outperform the methods we summarized below.

\paragraph{XGBoost enhanced with weighted embedding} 
XGBoost, as a scalable tree boosting system~\cite{Chen2016}, builds an ensemble of weak trees by incrementally adding new instances that contribute the most to the learning objectives.
To accommodate the distributed text representation in numeric format, we average the embedding of each word per message as proposed by~\citet{Stein2019} with some modifications. 
First, as we deal with medical conversations, we use Wikipedia PubMed as our word embedding representation. 
Second, since simple averaging does not reflect the importance of each word, we use a weighted average where TF-IDF values of the words are the weights. 
By these slight adaptations, we ensure that unimportant words do not have an impact on the averaged output for a given message~\cite{zhao2015}. 
Finally, we append the length of the patient message as a new independent feature. 
Hence, the text representation along with its length contributes to 201 independent attributes for each patient message.
\textcolor{black}{
We note that, in our preliminary experiments, we also adopted doc2vec in addition to word2vec. 
However, the performance of doc2vec on our medical chats dataset was significantly lower than that of word2vec. 
Previous studies also reported similar results, demonstrating the better performance of word2vec over doc2vec~\citep{Chen2021, Shao2018,hughes2017medical, Zhu2016Word2vec}
Accordingly, we proceed with the weighted word2vec embeddings in our numerical study.
}For XGBoost, while we include the message length in the triggering phase, we exclude it in the response generation phase. 
In our preliminary analysis we compared this approach with both simple TF-IDF representation~\cite{Qi2020} and unweighted word embedding average \textcolor{black}{and found 5.7\% and 2.9\% improvement in precision@1, respectively.}

\paragraph{SVM enhanced with weighted embedding}
Support Vector Machine (SVM) has been widely used for text categorization and classification in different domains~\cite{Hartmann2019, Serkan2011, Ren2013}. 
It identifies support vectors (i.e., data points closer to the hyperplane) to position a hyperplane that maximizes the classifier's margin.
SVM learns independently the dimensionality of the feature space, which eliminates the need for feature selection. 
It typically performs well for text classification tasks with less computational effort and hyperparameter tuning while also being less prone to overfitting~\cite{Joachims1998}. 
Hence, we consider SVM as a baseline, \textcolor{black}{and employ the same weighted embedding described for XGBoost in SVM training and testing}.

\paragraph{Bi-directional LSTM enhanced with Wikipedia PubMed embedding}
Long short-term memory (LSTM) units, as the name suggests, capture both the long-term and the short-term information through the input, forget, and output gates. 
Therefore, it has the ability to forget uncorrelated information while passing the relevant ones~\cite{Du2020}. 
Since the patient messages consist of long sentences, such gates are ideal to have the least information loss. 
They can detect message contents stored as the long-term memory inside the cell while keeping invaluable information provided towards the end of a sentence. 
In our algorithm, we use Bi-directional LSTM (BiLSTM) units that learn information from both directions, enabling them to access both the preceding and succeeding contexts~\cite{Liu2019}. 
This way, equal weight is provided towards the beginning and the end of a sentence. 
BiLSTM units are an appropriate remedy for our problem since a patient message may contain useful information either at the beginning of a sentence or at the end. 
Using the attention mechanism, BiLSTM disregards generic comments and concentrates on more pertinent information~\cite{Sachan2019}.

\paragraph{Seq2Seq enhanced with Wikipedia PubMed embedding} Sequence-to-sequence models turn one sequence into another. 
It is used primarily in text translation, caption generation, and conversational models. 
Therefore, we only apply it to the reply suggestion phase as it is not generalizable to the triggering phase. 
Our Seq2Seq model consists of an encoder, decoder, and an attention layer~\citep{NIPS2014}. 
The encoder encodes the complete information of a patient message into a context vector, which is then passed on to the decoder to produce an output sequence of a doctor's reply. 
Since our data consists of long sentences, we use an attention mechanism to assign more weight to relevant parts of the context vector to improve computational efficiency as well as accuracy~\cite{luong2015}. 

To prepare our data for the Seq2Seq model, we tokenize both doctors' and patients' messages and pad them to match the length of the longest sentence in our data. 
Start and end tokens are added to each sequence. 
Furthermore, we use a pre-trained Wikipedia PubMed embedding layer to capture the text semantics. 
The encoder additionally uses a Bi-directional LSTM layer for enhanced learning of encoded patient messages. 
We train it using the Adamax optimizer and sparse categorical cross-entropy to calculate the losses. 

We employ beam search~\citep{kumar2013beam} to retrieve the predicted outcomes of the model using a beamwidth of three and apply length normalization to avoid biases against lengthier doctor replies.
We rank replies according to their beam scores and choose the top-$k$ responses. 
As the model generates responses word by word, there is a tendency for the model to suggest inappropriate or grammatically incorrect sentences. 
To overcome this issue, we apply cosine similarity to match the generated responses with our canned response set and select the ones with the highest cosine similarity score. 
Hence, we ensure that the proposed options will have proper word choice and grammar. 
Nevertheless, when the final top-$k$ suggested replies overlap, we iteratively cycle through their cosine scores and pick the next best response until we reach $k$ unique suggestions.

\paragraph{Bidirectional Encoder Representations from Transformers} Bidirectional Encoder Representations from Transformers, or BERT in short, are replacing the state-of-the-art models across the NLP world. This cutting-edge technology involves pre-trained deep bidirectional representations conditioned on the left and right context in all layers~\cite{devlin2018bert}. Accordingly, it can be fine-tuned simply by adding an output layer, depending on the task. BERT has a self-attention mechanism that encodes text pairs (i.e., questions and answers) and treats them as a concatenated text pairs with bidirectional cross attention. It is highly efficient as the fine-tuning is relatively inexpensive and fast~\cite{devlin2018bert}.

\textcolor{black}{The core innovative part of the BERT architecture is the self-attention mechanism, which is a typical example of contextual word embeddings. 
Early applications of contextual embeddings include ULMFiT and ELMo.
However, transformer-based language models such as BERT have gained significant popularity in recent studies and have been shown to achieve state-of-the-art performance for many natural language processing tasks. 
ELMo can generate context-sensitive embeddings for each word in a phrase subsequently sent to downstream activities. 
BERT, on the other hand, employs a fine-tuning strategy that allows the entire language model to be adapted to a downstream job, resulting in a task-specific architecture that is able to capture the semantics of the text. 
By learning bidirectional representations using the Masked Language Modeling objective, BERT overcomes various limitations of ELMo, ULMFiT, and GPT~\citep{Katikapalli2021}.
Recent studies showed the BERT-based models outperform ELMo and ULMFiT across different languages~\citep{Ameri2021, Bataa2019, Fatima2021, wang2020}, showing the strength of BERT as a contextual word embedding.
To the best of our knowledge, this is the first study where transformers are employed to facilitate patient-doctor chats.
}

\textcolor{black}{
We leverage BERT in both the triggering and response generation phases. For our experiments, we utilize a BERT base architecture that was pretrained on MEDLINE/PubMed dataset\footnote{~\href{https://tfhub.dev/google/experts/bert/pubmed/2}{https://tfhub.dev/google/experts/bert/pubmed/2}}. This pretrained model is aimed to be used in the medical domain and is suitable for our experiments. 
In BERT model training, we use an Adam optimizer and a batch size of 16. 
In addition, we employ early stopping during the training phase to avoid overfitting.
Early stopping patience of 5 is chosen for mitigating local optimality, while the number of epochs is set to 50. 
We fine-tune the large pretrained language models by attaching a dropout layer with parameter 0.1. 
The weights of the best-performing model on the validation set are stored and utilized in the testing phase. 
}

\subsubsection{Rule-based baselines} \label{sec:rule-based}
\textcolor{black}{
We consider three different approaches for trigger/response generation which we use as baselines in our comparative analysis with different machine learning models. 
Two of these methods are based on sequence (i.e., text) similarity, and the third one generates trigger and response outcomes randomly according to the frequency of the labels in the training set.
}

\paragraph{TF-IDF Similarity} 
\textcolor{black}{
In this approach, we first generate a TF-IDF matrix of the vocabularies in each patient message. 
Next, we compute the cosine similarity between a given test instance and all the training instances using the corresponding TF-IDF values.
We select the suggested binary triggering value or the doctor response based on the most similar instance in the training set. 
As such, this approach generates a triggering prediction and response generation for any given test instance solely based on text similarity without employing any sophisticated model. 
}

\paragraph{Weighted TF-IDF Similarity}
\textcolor{black}{
TF-IDF Similarity method only utilizes the vocabulary frequencies, and it does not take into account the semantic similarity of the words.
For instance, ``disease'' and ``illness'' are typically treated as completely different words when using TF-IDF. 
To alleviate this issue, we extract the word embedding of each vocabulary from the Wikipedia PubMed word embedding representation. 
Then, we use a weighted average of the word embedding of vocabularies in each sentence, with the weights being the corresponding TF-IDF values. 
As such, we ensure that insignificant words do not influence the average output for a particular message.
After generating a weighted embedding matrix, we employ cosine similarity and associate each instance in the test set with its most similar counterpart in the training set.
}

\paragraph{Frequency-based}
\textcolor{black}{
The frequency baseline ranks possible responses in the training corpus in order of their frequency, thus encapsulating the extent to which we can recommend these frequent responses regardless of the original messages' content. 
Specifically, this baseline chooses the responses according to their probability distributions in the training set. 
For instance, if 80\% of the patient messages have an associated response, then, in the triggering phase, this approach returns 1 (i.e., ``trigger response'') with a probability of 0.8, and it gives 0 (i.e., ``do not trigger response'') with a probability of 0.2.
Similarly, in the response generation phase, this approach returns random responses according to their probability distributions.
}

\subsection{Experimental setup}\label{sec:experimntal_setup}
We use 5-fold nested cross-validation for train-test split and tune the most important parameters of the models by dividing the training dataset into validation and training sets. Accordingly, one fold is used for testing, and the other four folds are divided into validation and training sets based on an 80-20 split. Each fold has approximately 6,281 patient-doctor message pairs. 
In each grid search procedure of the hyperparameters, we identify the best models for text classification. 
In the testing phase, we use the models that perform best on the validation set. We provide the final model configurations as follows.
\begin{itemize}
    \item We obtain the learning rate as 0.3 and the number of trees as 200 for XGBoost, whereas the other parameters are set to the default values of the XGBoost library in Python. 
    
    \item Our SVM model uses a linear kernel with a degree of 3 and is implemented using the scikit-learn package in Python.
    
    \item We use TensorFlow to create our sequential LSTM model, which has an embedding layer powered by Wikipedia Pubmed. The model has a bidirectional layer of size 200, a dense layer of size 100 with the Relu activation, and a dense output layer with sigmoid and softmax activation functions for triggering and response generation, respectively. 
    We train the model for 20 epochs using the Adam optimizer. 
    
    \item For our Seq2Seq model, we initiate our encoder with the Wikipedia Pubmed embedding layer, followed by a bidirectional LSTM layer of size 1024. 
    Next, we use an LSTM layer for our decoder, along with the Luong Attention Mechanism. We calculate sparse categorical cross-entropy loss, and the model is trained for 15 epochs using the Adamax optimizer.  
    
    \item For our BERT model, we apply a BERT processor and encode it using a vocabulary for English extracted from Wikipedia and BooksCorpus and fine-tuned on PubMed abstracts~\citep{gu2020domain}. Afterwards, we apply a 10 percent dropout and a dense output layer with sigmoid and softmax activation functions for triggering and response generation, respectively. 
\end{itemize}

\subsection{Performance metrics}
To investigate the performance of two different phases of the algorithm, we relied on two different sets of metrics. 
For the triggering phase, we used ``accuracy'', the ratio of correct predictions to the number of instances, ``precision'', how many instances predicted as class $c$ belongs to the same class, ``recall'', how many data points that belong to class $c$ are found correctly, and ``F1-score'', which is the harmonic mean of precision and recall. 
These four performance metrics are threshold-dependent (i.e., model predictions constitute a probability distribution over class labels, and binary predictions are determined based on a probability threshold, e.g., 0.5).
Therefore, we also utilize the area under the ROC curve (AUC-ROC) as a threshold-independent approach to mitigate the problems with threshold settings~\cite{Hern2012}.

We employ different metrics to assess the performance of response generation models. 
We mainly rely on the ``precision@k'' metric to report the accuracy of the suggestions~\citep{kannan2016smart}.
If the suggested response is among the top $k$ responses, we recognize it as a correct suggestion; otherwise, it is regarded as an unsuitable suggestion. 
We take the number of generated responses as $k=3$. 
Therefore, if the model is adept enough to include the proper reply among the top 3, it will be considered an appropriate suggestion. 
We also report ``precision@1'' and ``precision@5'' to gain more insights regarding the models' performances. 
Another useful metric is the rank of the suggested response. 
If a model puts forward a reply in rank 4, there is a likelihood that it can be improved further by some parameter tuning. 
On the other hand, if the response is ranked 20th, the model is unlikely to suggest a proper response. 
Consequently, we report the Mean Reciprocal Rank (MRR) metric for the proposed responses, that is, 
\begin{equation*}
    \text{MRR} = \frac{1}{N}\sum_{i=1}^{N}{\frac{1}{\text{rank}_i}}
\end{equation*}
where $N$ is the total number of messages. 
MRR ranges from 0 to 1, where 1 indicates the optimal performance (all the suggestions are ranked first). 

\section{Results}\label{sec:results}
Our model has two phases, namely, triggering and response generation. In this section, we first report the performance of the triggering filter and then the smart response suggestion. We discuss the model performance for each phase in detail and provide sample generated responses. Finally, we investigate the sensitivity of the model to the triggering threshold.

\subsection{Triggering performance}
Accurate triggering is important since an infeasible patient's message passing this filter not only leads to an irrelevant message but also increases the computational complexity.
Consequently, an inferior model will reduce the quality of the suggested replies and degrade the performance of the overall response generation mechanism.

Table~\ref{tab:triggering_acc} shows the performance of different models for the triggering phase. 
All the models outperform the baseline approaches, which generates the response based on either their frequencies or similarities. 
The reason for the relatively high accuracy of the frequency-based approach is the imbalanced ratio of feasible and infeasible cases. 
Therefore, by overestimating the majority group, it still can reach acceptable performance. 
However, when it comes to the threshold-independent metric, AUC-ROC, the frequency-based approach performs as poorly as a random guess with almost 50\% AUC-ROC. 
On the other hand, other approaches show significant improvement over the TF-IDF similarity and weighted TF-IDF similarity baselines. 

\setlength{\tabcolsep}{4pt} 
\renewcommand{\arraystretch}{1.3} 
\begin{table}[!ht]
\centering
\caption{\textcolor{black}{Performance of different models for the triggering task (reported values are all in percentages)}}\label{tab:triggering_acc}
\resizebox{\textwidth}{!}{
\begin{tabular}{lrrrrrrrr}
 \toprule
\multirow{2}{*}{\textbf{Method}} & \multirow{2}{*}{\textbf{Accuracy}} & \multicolumn{2}{c}{\textbf{Precision}} & \multicolumn{2}{c}{\textbf{Recall}} & \multicolumn{2}{c}{\textbf{F1-score}} & \multirow{2}{*}{\textbf{AUC-ROC}} \\
 &  & Infeasible & Feasible & Infeasible & Feasible & Infeasible & Feasible &  \\

\midrule
\rowcolor[HTML]{EFEFEF} 
\textbf{BERT\textsuperscript{\textdaggerdbl}} & \textbf{87.82$\pm$0.6} & \textbf{80.51$\pm$2.8} & 89.59$\pm$1.8 & 62.70$\pm$7.7 & \textbf{95.32$\pm$1.5} & 70.08$\pm$3.6 & \textbf{92.34$\pm$0.3} & \textbf{93.93$\pm$0.2} \\
\textbf{BiLSTM\textsuperscript{\textdagger}} & 87.37$\pm$0.4 & 72.36$\pm$2.9 & \textbf{92.14$\pm$1.4} & \textbf{73.72$\pm$5.5} & 91.44$\pm$1.9 & \textbf{72.81$\pm$1.4} & 91.76$\pm$0.4 & 91.71$\pm$0.5 \\
\textbf{XGBoost\textsuperscript{\textdagger}\textsuperscript{*}} & 86.03$\pm$0.3 & 73.78$\pm$1.7 & 88.93$\pm$0.2 & 61.10$\pm$0.5 & 93.49$\pm$0.6 & 66.83$\pm$0.5 & 91.15$\pm$0.2 & 92.38$\pm$0.3 \\
\textbf{SVM\textsuperscript{\textdagger}\textsuperscript{*}} & 85.68$\pm$0.3 & 74.93$\pm$2.2 & 87.99$\pm$0.3 & 56.98$\pm$1.4 & 94.27$\pm$0.8 & 64.69$\pm$0.5 & 91.02$\pm$0.2 & 92.21$\pm$0.5 \\
\hdashline
\textbf{Weighted TF-IDF\textsuperscript{\textdagger}} & 75.36$\pm$0.8 & 47.54$\pm$1.3 & 88.68$\pm$0.2 & 66.73$\pm$0.6 & 77.94$\pm$1.1 & 55.51$\pm$0.7 & 82.96$\pm$0.6 & 72.34$\pm$0.4 \\
\textbf{TF-IDF} & 75.43$\pm$0.6 & 47.39$\pm$1.1 & 87.04$\pm$0.1 & 60.16$\pm$0.7 & 80.00$\pm$1.0 & 53.00$\pm$0.4 & 83.37$\pm$0.5 & 70.08$\pm$0.2 \\
\textbf{Frequency} & 64.74$\pm$0.3 & 23.21$\pm$0.9 & 77.03$\pm$0.3 & 23.03$\pm$1.2 & 77.21$\pm$0.6 & 23.11$\pm$1.0 & 77.12$\pm$0.2 & 49.99$\pm$0.7 \\
\bottomrule
\multicolumn{9}{l}{\quad \textsuperscript{\textdaggerdbl}: PubMedBERT embedding; \quad \textsuperscript{\textdagger}: Wikipedia-PubMed embedding; \quad \textsuperscript{*}:TF-IDF values as weight}\\
\end{tabular}
}
\end{table}

Table~\ref{tab:triggering_acc} also indicates that BERT, BiLSTM, XGBoost, and SVM perform similarly in predicting the majority class. 
All the models have acceptable performance in terms of precision, recall, and F1-score for the feasible cases. 
However, XGBoost and SVM perform relatively poorly in recalling the infeasible messages. 
They overfit the majority class, leading to passing a high number of messages to the next phase and, consequently, an abundance of irrelevant reply suggestions. 
BERT model that uses PubMedBERT  shows the best aggregate performance over various performance metrics.
Nonetheless, SVM and XGBoost demonstrate slight superiority in some metrics, which mainly stems from overestimating one class and neglecting the other one. 
Therefore, we choose BERT as our primary triggering model. \textcolor{black}{
We also explore the best-performing triggering model when combined with the response suggestion model in Appendix~\ref{sec:best_pipline}.}

\subsection{Response suggestion performance}
After a message successfully passes the triggering filter, it enters into the response suggestion model. 
Response suggestion aims to suggest proper messages within the top responses to facilitate the patient-doctor conversation.
Here, we only concentrate on doctors' response generation processes. 

We compare machine learning algorithms with the baseline, frequency-based suggestion. 
The baseline selects doctor responses from the canned messages based on their occurrence probability in the training set. 
As our response set includes certain frequent categories, the precision of the baseline might seem relatively high. 
However, the difference between machine learning algorithms and the baseline is statistically significant.
Table~\ref{tab:response_acc} summarizes the accuracy of each algorithm. 
BERT enhanced by PubMedBERT embedding significantly outperforms other machine learning and deep learning algorithms.
It suggests more correct responses in the first rank (precision@1) than other alternative approaches. 
In 59.32\% of the suggestions, the proper reply does exist in its first suggestion, and in 85.42\% of the times, the model can generate a response desired by a doctor (i.e., in top-3 responses). 
In a software application that suggests top-3 responses, the model accuracy based on precision@3 is promising. Table~\ref{tab:response_acc} also reports the Mean Reciprocal Rank of the models. According to the MRR values, BERT significantly outperforms others. It is important to note that the model is able to generate an instantaneous response, i.e., less than a second, which is desired by the application. Therefore, the proposed model can suggest the proper reply options in a timely manner.

\setlength{\tabcolsep}{6pt} 
\renewcommand{\arraystretch}{1.3} 
\begin{table}[!ht]
\centering
\caption{\textcolor{black}{Accuracy of the suggested responses for different models}}\label{tab:response_acc}
\resizebox{.9\textwidth}{!}{
\begin{tabular}{lrrrrr}
\toprule
 \textbf{Method} & \textbf{precision@1 (\%)} & \textbf{precision@3 (\%)} & \textbf{precision@5 (\%)} & \textbf{MRR} 
 \\
\midrule
\rowcolor[HTML]{EFEFEF} 
\textbf{BERT\textsuperscript{\textdaggerdbl}} & \textbf{59.32$\pm$1.64} & \textbf{85.42$\pm$0.82} & \textbf{87.56$\pm$0.63} & \textbf{0.79$\pm$0.00} \\
\textbf{BiLSTM\textsuperscript{\textdagger}} & 58.98$\pm$0.88 & 83.28$\pm$0.75 & 85.37$\pm$0.62 & 0.75$\pm$0.00 \\
\textbf{Seq2Seq\textsuperscript{\textdagger}} & 53.21$\pm$0.61  & 61.48$\pm$4.46 & 68.63$\pm$5.54 & 0.60$\pm$0.02 \\
\textbf{XGBoost\textsuperscript{\textdagger}\textsuperscript{*}} & 51.33$\pm$0.31 & 79.59$\pm$0.52 & 82.83$\pm$0.42 & 0.69$\pm$0.01 \\
\textbf{SVM\textsuperscript{\textdagger}\textsuperscript{*}} & 47.62$\pm$0.42  & 78.97$\pm$0.49 & 82.25$\pm$0.51 & 0.68$\pm$0.00 \\
\hdashline
\textbf{Weighted TF-IDF} & 34.41$\pm$2.57 & 46.46$\pm$2.97 & 53.53$\pm$1.2 & 0.42$\pm$0.03 \\
\textbf{TF-IDF} & 32.32$\pm$2.11 & 44.35$\pm$1.28 & 51.01$\pm$0.68 & 0.42$\pm$0.03 \\
\textbf{Frequency} & 16.71$\pm$0.25 & 32.20$\pm$0.67 & 42.80$\pm$0.74 & 0.35$\pm$0.00 \\
\bottomrule
\multicolumn{5}{l}{\quad \textsuperscript{\textdaggerdbl}: PubMedBERT embedding; \quad \textsuperscript{\textdagger}: Wikipedia-PubMed embedding; \quad \textsuperscript{*}:TF-IDF}\\
\end{tabular}
}
\end{table}


\textcolor{black}{The high performance of our adopted algorithm within the initial ranks (e.g., for $k=1$ and $k=3$) implies that the model is able to derive deep insights from the textual information present in a patient message, which enables capturing the details beyond a basic understanding of the message intent. 
Such ability to learn the semantics of the text can be particularly impactful for the messages that consist of highly overlapping themes and ideas.
Significant performance gain over baseline approaches also points to the ability of the machine learning models to capture the semantics of the patient messages.
}

\begin{figure}[!ht] 
    \centering
    \includegraphics[width = 0.7 \linewidth]{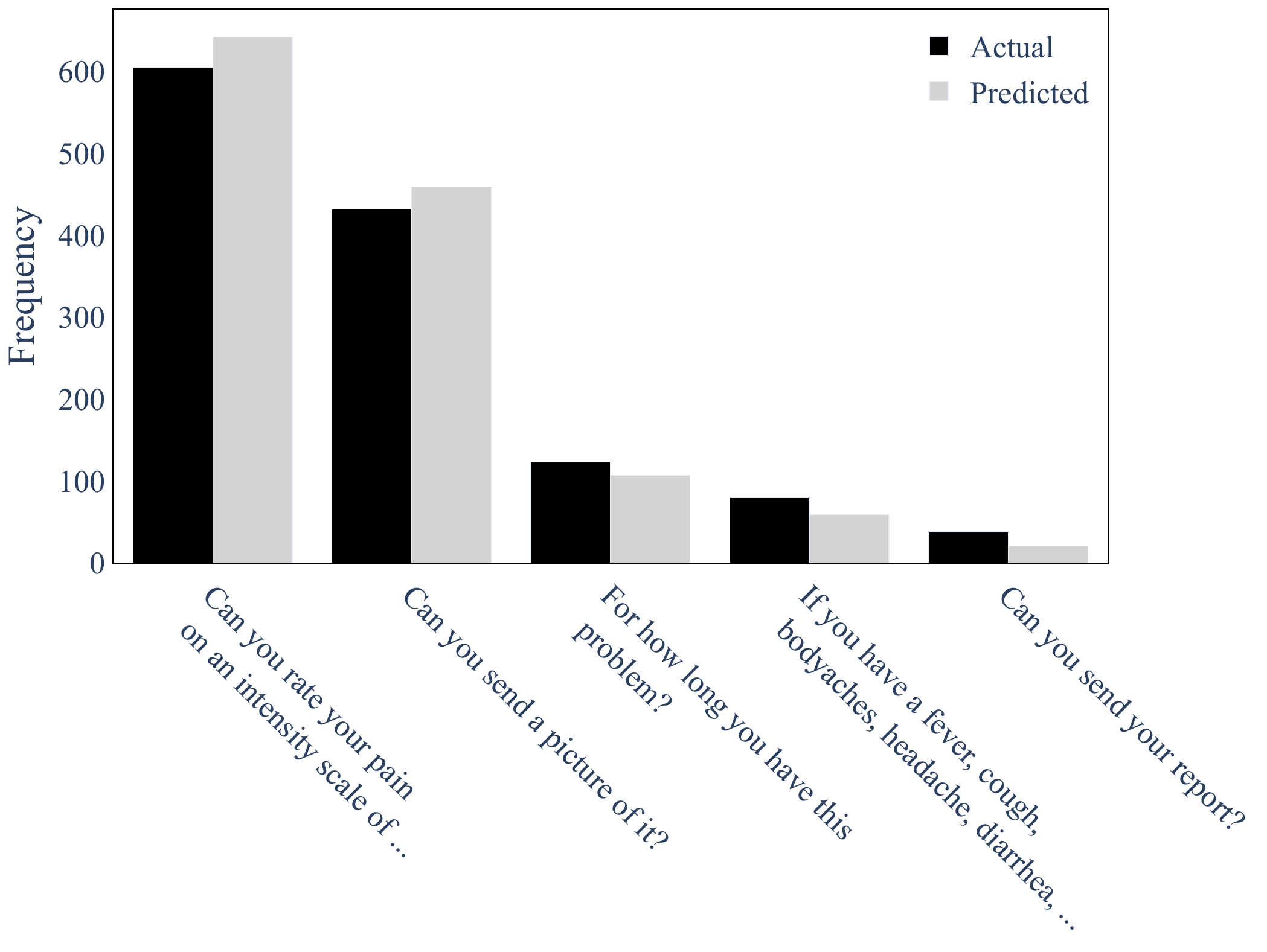}
    \caption{The distribution of the actual frequency of the doctor messages compared to the predicted responses by the final model }
    \label{fig:response_freq}
\end{figure}

We demonstrate the distribution of the actual frequency of the most frequent medical responses (i.e., excluding casual responses such as ``You are welcome.'' and ``Thanks.'') in Figure~\ref{fig:response_freq}. 
The ground-truth frequency is shown in black, while the predicted frequency is in grey. 
We observe that both the prediction and the ground truth follow a similar distribution. 
For the casual responses, which are excluded from the graph, the generated frequencies exhibit a similar pattern as the actual ones.   

\begin{figure}[!ht] 
    \centering
    \includegraphics[width = 0.9 \linewidth]{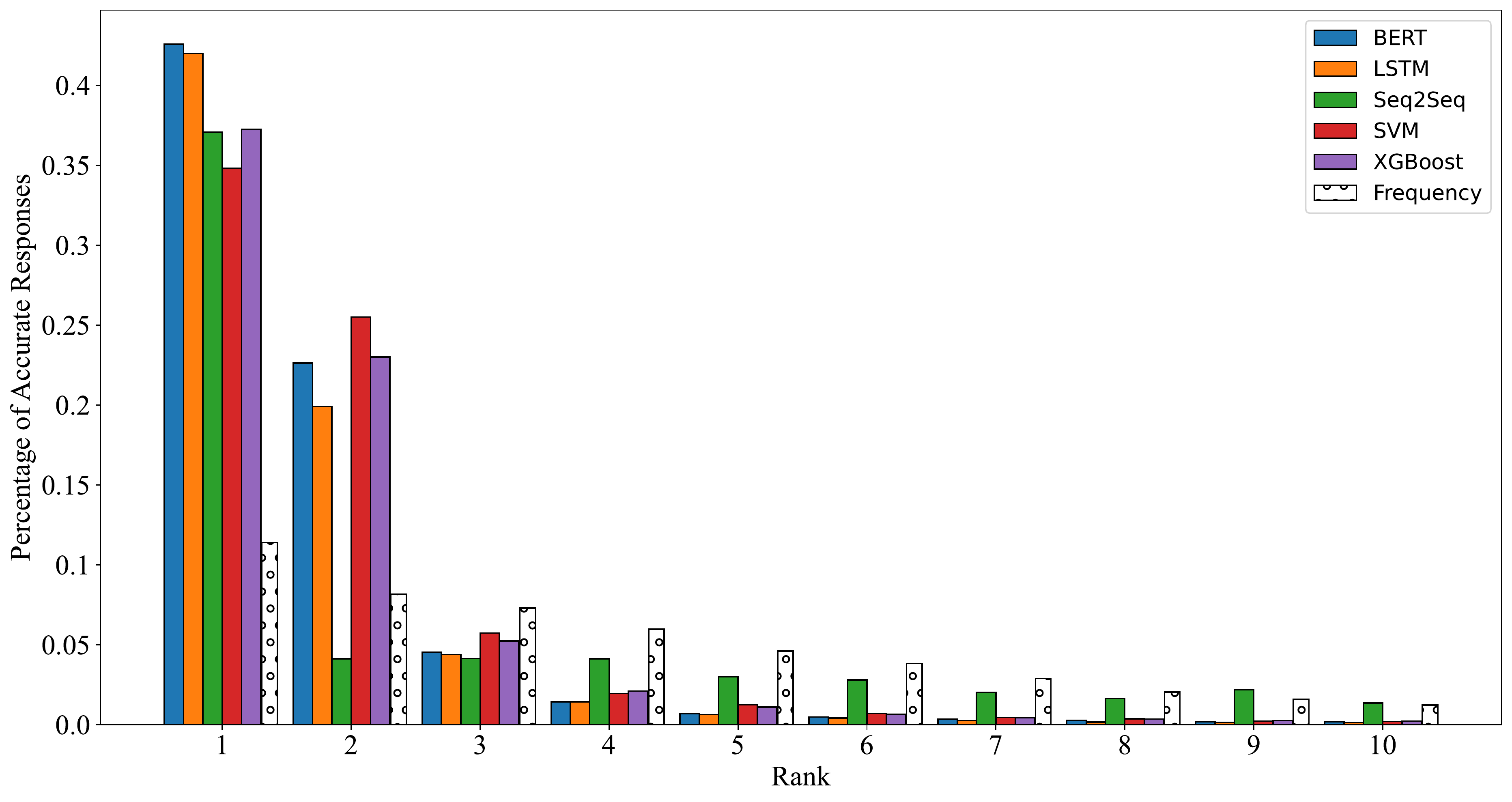}
    \caption{The distribution of correct responses ranked top-10 for different methods}
    \label{fig:rank_response}
\end{figure}

Figure~\ref{fig:rank_response} shows the accuracy of suggested responses given their rank. 
If a message passes the triggering phase, the model will recommend a reply whether the original message is feasible or not. 
Therefore, the plot shows the ratio of precise suggestions per rank for all feasible and infeasible cases entering the response generator. 
BERT has the best overall performance for the top-3 responses.
It prompts the highest ratio of correct replies in the first rank while having the least suggestions in the fourth position and above. 
Surprisingly, Seq2Seq, which is the third-best algorithm considering the suggested messages ranked first, loses its superiority shortly after.
One reason for the performance drop is the beam search associated with the response selection. 
It does not have the option to diversify the message, and adding rule-based diversification increases its computation time.
Therefore, the model fails to generate high-quality responses considering the pace needed to output a reply. 
All the analyses highlight the advantage of using BERT in automated doctor response recommendations. 

\subsection{Sample generated responses}
\begin{figure}[!ht]
     \centering
     \begin{subfigure}[b]{0.4\textwidth}
         \includegraphics[width=\textwidth]{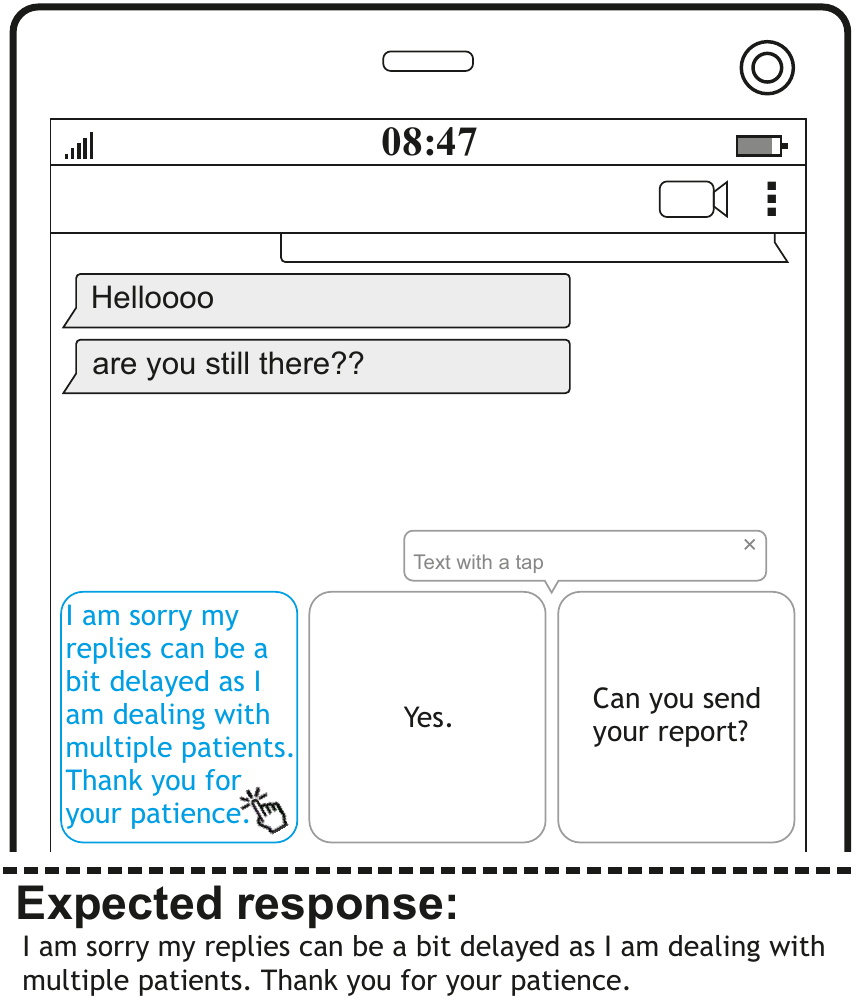}
         \caption{Example 1: A response appears at the first position as expected.}
         \label{fig:smart_response1}
     \end{subfigure}
     \hspace{1cm}
     \begin{subfigure}[b]{0.4\textwidth}
         \includegraphics[width=\textwidth]{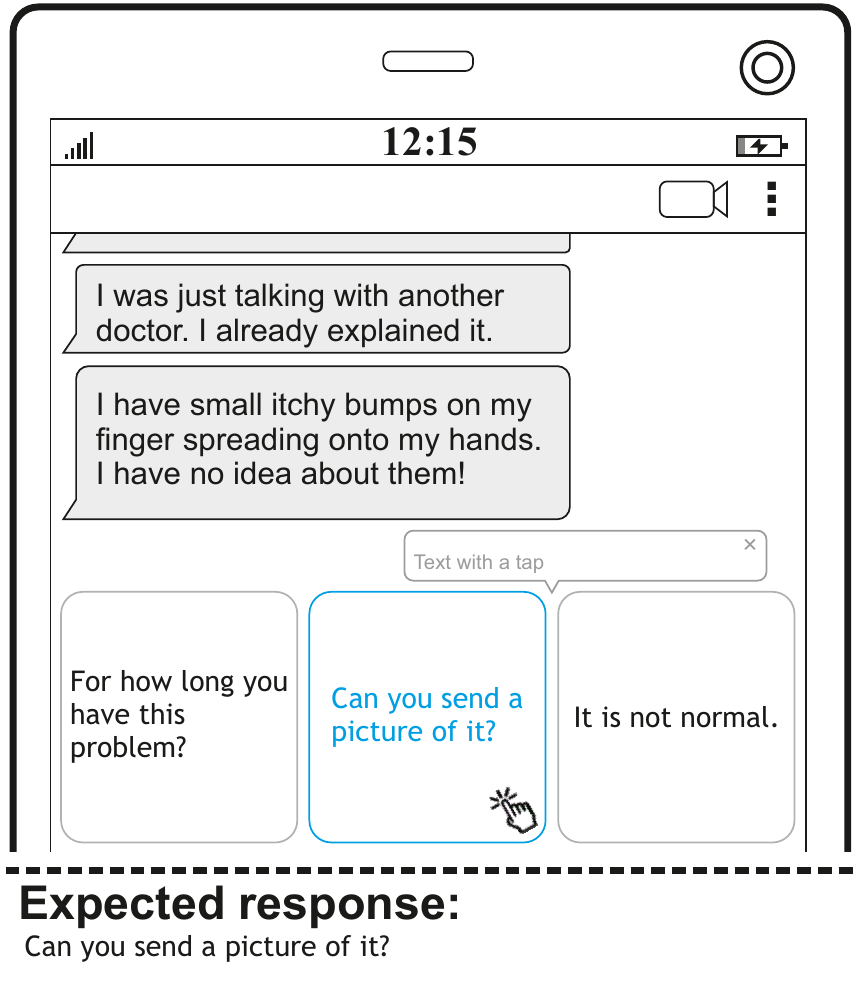}
         \caption{Example 2: The expected response is still among the top-3 suggestions.}
         \label{fig:smart_response2}
     \end{subfigure} 
     
     \bigskip
     \begin{subfigure}[b]{0.4\textwidth}
         \includegraphics[width=\textwidth]{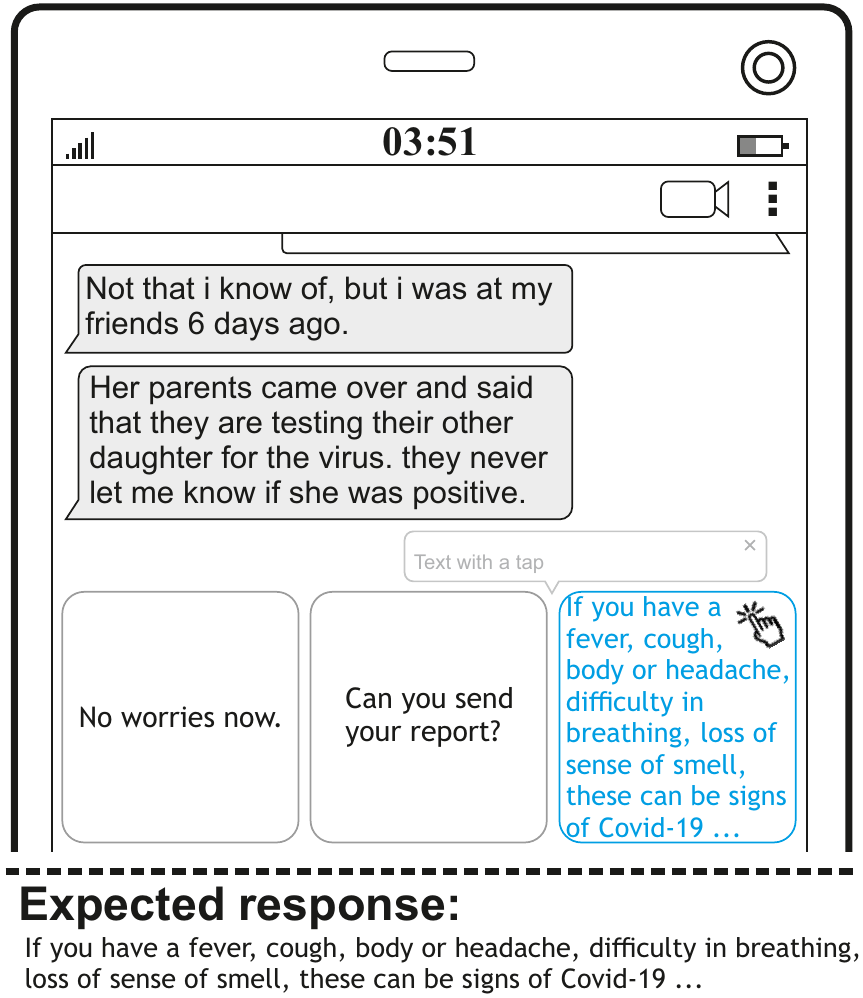}
         \caption{Example 3: The expected reply is ranked third. Still easy to be tapped.}
         \label{fig:smart_response3}
     \end{subfigure}
     \hspace{1cm}
     \begin{subfigure}[b]{0.4\textwidth}
         \includegraphics[width=\textwidth]{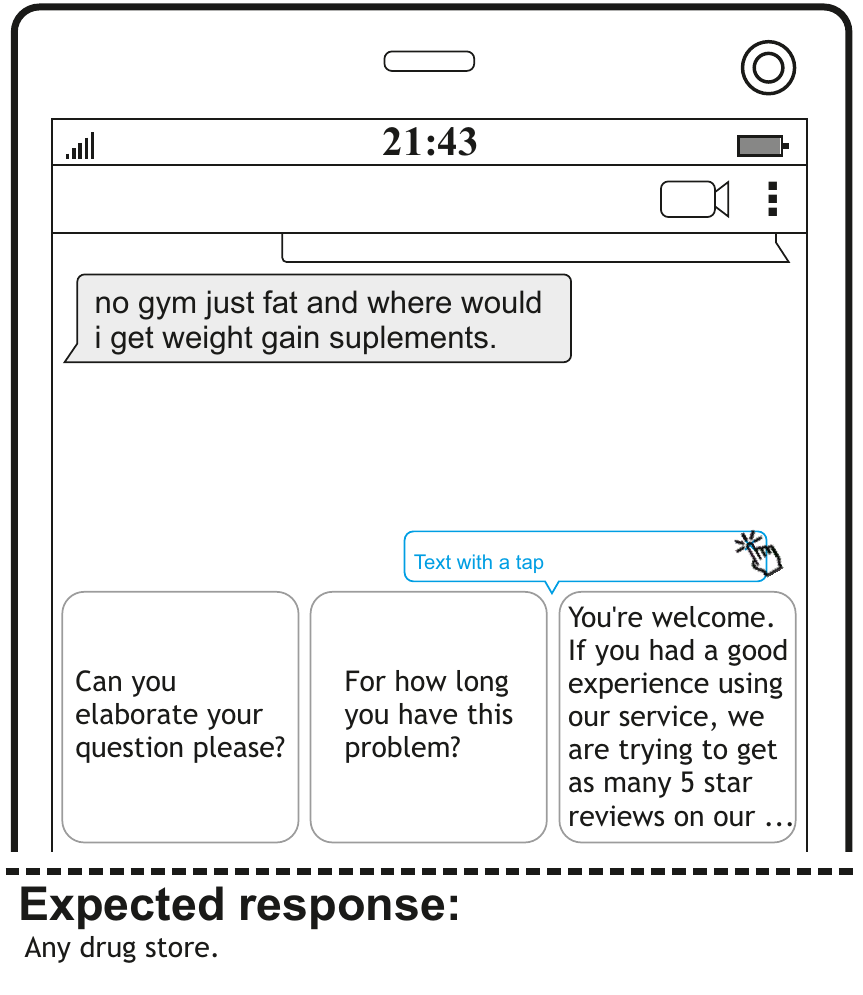}
         \caption{Example 4: A wrong suggestion. The model does not suggest the expected reply.}
         \label{fig:smart_response4}
     \end{subfigure}     
        \caption{Examples of smart responses generated by the final model}
        \label{fig:smart_response}
\end{figure}

Figure~\ref{fig:smart_response} demonstrates sample replies generated by the final model. 
If a patient's message passes the triggering phase, the algorithm suggests top-3 possible responses in order. 
In Figure~\ref{fig:smart_response1}, the proper response is ranked first. 
The model learned from the training set to select an automated message from the canned set to ``apologize for delays''; otherwise, it is confusing. 
As these kinds of less frequent messages are not diversified, and we do not have other alternatives for them, the third suggested response might seem less relevant.
Figure~\ref{fig:smart_response2} depicts a situation in which the expected response is ranked second. 
However, the other two responses are pertinent to the patient's message. 
As the patient asks about itchy bumps, the doctor may either request a picture to have a better idea or ask about its longevity. 
In Figure~\ref{fig:smart_response3}, the patient implies Covid-19 condition by referring to ``virus'', ``testing'', and ``positive'', and the suggested response inform the patient about the symptoms of Covid-19. 
The other two responses show compassion for the patient and ask for a report if there is any clinical test available. 
Although the response does not appear at the first rank, others still provide an appropriate alternative for the ground truth.
Lastly, Figure~\ref{fig:smart_response4} shows a case in which the proper response is not provided. 
The patient asks for the place where he/she can find weight gain supplements, and the model cannot suggest the location. We manually checked for the rank of the correct reply and found it at rank six. Nonetheless, there is a typo in the patient's message that can be addressed, and the wrong responses can be potentially avoided.

\subsection{Sensitivity of the model to the triggering threshold} \label{sec:sensitivity}
One of the most significant parameters of the algorithm is the triggering threshold. 
As the triggering model suggests the probability of generating a response, it is important to determine how to convert that probability to a binary decision. 
As a rule of thumb, we round numbers greater than or equal to 0.5 to 1 and smaller ones to 0. 
However, the question is whether the threshold of 0.5 provides the best-suggested replies. 
When the threshold is too small, the model tends to generate responses for most infeasible cases; on the other hand, when it is close to 1, the model becomes more conservative as it avoids generating inappropriate responses.
\begin{figure}[!ht]
     \centering
     \begin{subfigure}[b]{0.47\textwidth}
         \includegraphics[width=\textwidth]{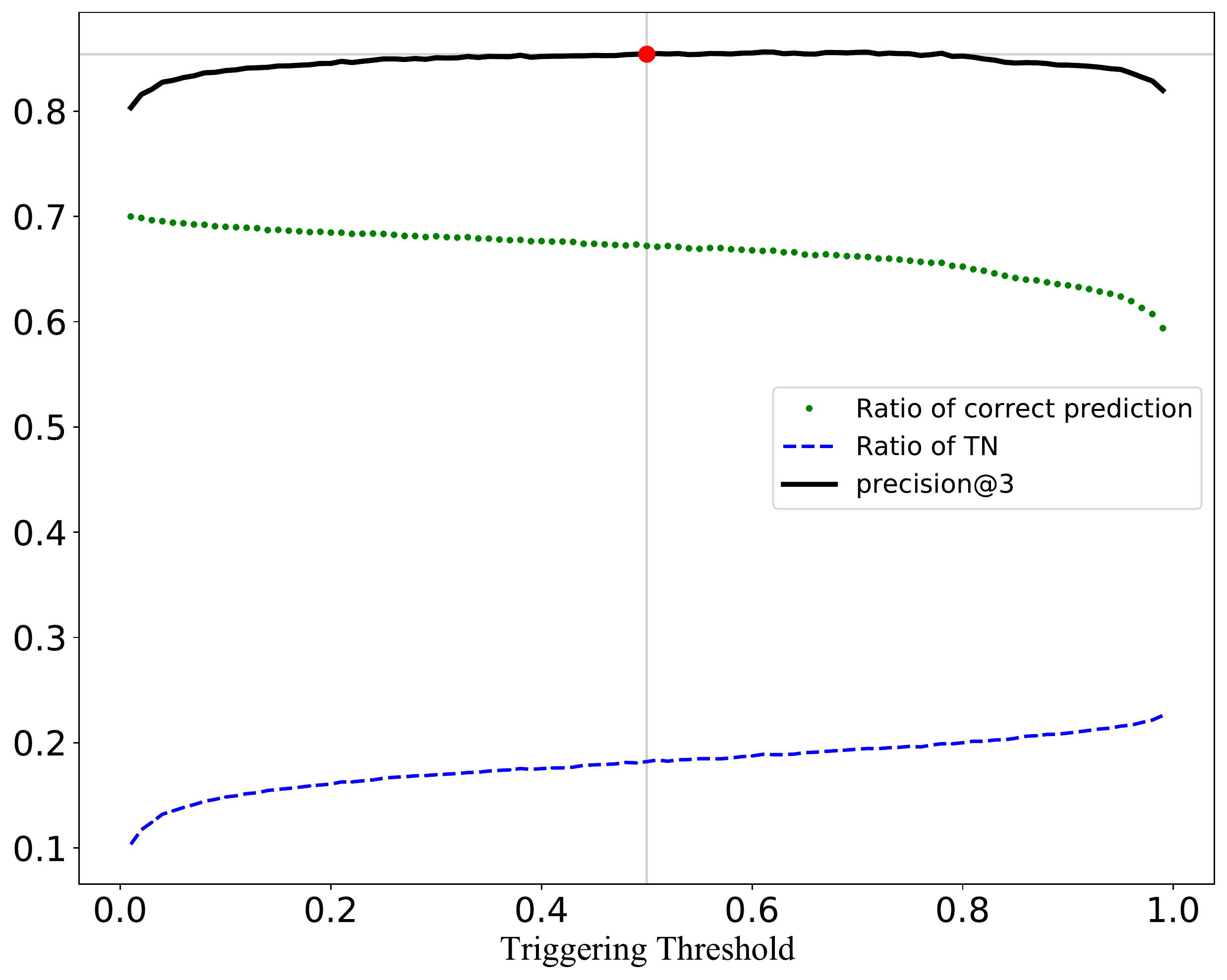}
         \caption{Correct predictions in detail}
         \label{fig:precision_at_3}
     \end{subfigure}
     \begin{subfigure}[b]{0.47\textwidth}
         \includegraphics[width=\textwidth]{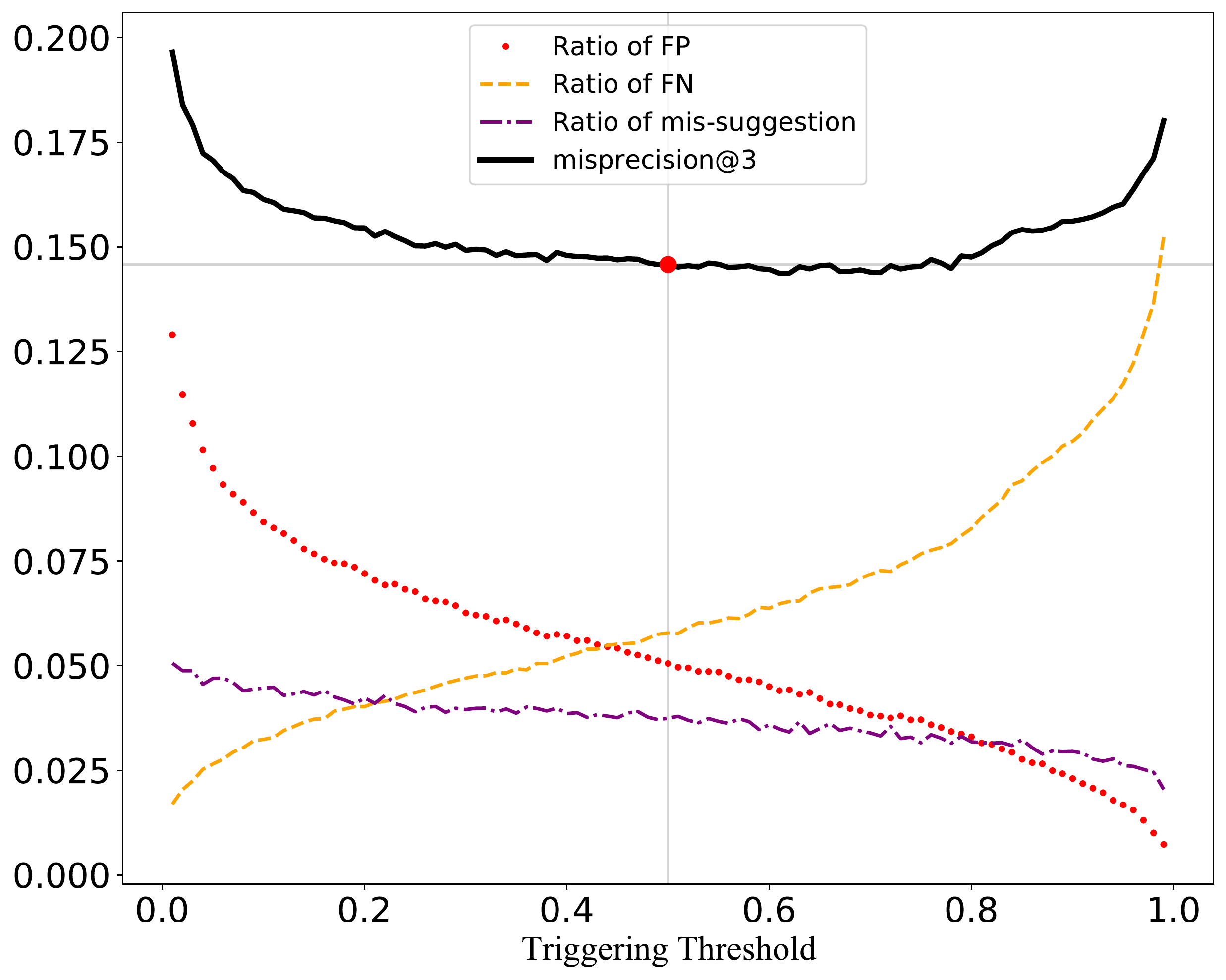}
         \caption{Wrong predictions in detail}
         \label{fig:misprecision_at_3}
     \end{subfigure} 
        \caption{The sensitivity of the algorithm to the threshold of the triggering phase}
        \label{fig:sensitivity}
\end{figure}

Figure~\ref{fig:sensitivity} demonstrate the sensitivity of the performance of the BERT model to its triggering threshold.
Figure~\ref{fig:precision_at_3} demonstrates the source of precision. 
It can come either from the correct filtering of infeasible responses (TN) or the correct suggested replies that fall in the top-3 suggestions. 
Therefore, the total precision@3, the sum of the other two predictions, is shown in black. 
We observe that the algorithm is robust to the threshold parameter as there is only a negligible fluctuation in the total precision for different values of the triggering threshold. 
In Figure~\ref{fig:misprecision_at_3}, we explore the effect of triggering threshold on the ratio of mispredictions. 
We divide the misprecision@3 into three folds: the incorrect predictions originating from passing infeasible messages (FP), the ratio of incorrectly filtered feasible messages (FN), and missuggestions of feasible responses, meaning a suggested response is not among the top 3. 
Same as the previous plot, we depict the sum of these three subdivisions in black. 
The misprecision@3 has the least effect on the ratio of missuggestions; however, it highly affects both False Positive and False Negative cases. 
Altogether, misprecision@3 does not oscillate within the range of 0.3 and 0.8. 
Hence, the method is found to be robust under different threshold values. Accordingly, we choose the common threshold of 0.5 for all the experiments.

\section{Concluding remarks}\label{sec:conclusion}
Considering the rapid growth of online medical chat services, telehealth companies may choose to either expand their capacity --- e.g., the number of physicians --- or facilitate the communication for the existing employees to maximize their utilization. 
Instead of employing an expensive workforce, the cost of enhancing the current application with Artificial Intelligence is almost negligible. 
Accordingly, smart response suggestions can relieve the doctors' burden by facilitating patient-doctor communication, proposing appropriate replies, and saving their valuable time. 
To the best of our knowledge, we investigate the feasibility of having smart reply suggestions in medical contexts for the first time. 

We use the actual conversations between patients and doctors coming from an online medical chat service.
Accordingly, after exploratory data analysis, we clean the dataset by devising a canned response set. 
Using clustering techniques, we find the densest clusters of doctors' messages and extract frequent responses from those.
Afterward, we match the patient and doctor messages being aware of the complexity of disorderly exchanged chats, which results in 31,407 paired messages. 
Not all patient messages require smart replies; therefore, we also label the pairs as ``feasible'' or ``infeasible''. 
Our algorithm proceeds in two steps: predicting whether we need to trigger a smart reply and suggesting the proper response given a message passes the triggering phase. 
We explore different combinations of machine learning and deep learning algorithms to address each step. 
Furthermore, we tune the parameter and report the performance using 5-fold nested cross-validation. 
We assess each algorithm's performance using threshold-dependent and -independent metrics and observe that BERT is the best method for the triggering phase. 
It has a balanced score for both majority and minority class labels, i.e., feasible and infeasible cases.
In addition, its suggested replies are also the most appropriate in the response generation phase. 
Moreover, we tested its robustness to the triggering threshold and found it to be resilient to its parameter changes. 

\textcolor{black}{
The procedures we outline in this paper can be used in other similar systems, e.g., EMR systems; however, the models need to be retrained on the new corpus from the new ecosystem. As such, the proposed smart reply mechanism can be adapted to similar systems by following our suggested steps.
} A relevant venue for future research would be to improve the method by including more data points (i.e., more labeled-conversations). To the best of our knowledge, there is no publicly available dataset for medical conversations. Therefore, we only apply the algorithm to our proprietary dataset.
Besides, in response to the COVID-19 pandemic, our dataset is continuously being updated. Specifically, we find constant changes in patient queries and doctor answers. 
For instance, with regards to the modeling symptoms' questions, we observe that the vaccine queries become dominant.
Accordingly, an automated mechanism to retrain the models according to unprecedented challenges can be developed. 
We note that the overall response generation mechanism becomes feasible by introducing enough paired messages and updating the model weights. 
Moreover, as manual labeling is a tedious task, we plan to investigate semi-supervised learning for semantic clustering and labeling big datasets. 

\section*{Acknowledgement}
The authors would like to thank Your Doctors Online for funding and supporting this research. This work was also funded and supported by Mitacs through the Mitacs Accelerate Program. The authors would also like to thank Gagandip Chane for his help with the data labeling.

\section*{Conflict of Interest Statement}
None of the authors of this paper has a personal or financial relationship with people or organizations that could inappropriately influence or bias the result of the manuscript.

\bibliographystyle{elsarticle-num-names}

\singlespacing
\bibliography{references}

\newpage
\appendix
\section{\textcolor{black}{Finding the best pipeline}}\label{sec:best_pipline}
\textcolor{black}{
In this section, we discuss the evaluation results of the end-to-end pipelines obtained by combining different models for triggering and response generation. 
Table~\ref{tab:best_pipline} demonstrates the performance of various combinations of the triggering and response generation models. 
We use Precision@3 as the representative performance metric as we consider top-3 proposed responses in a generic chat application. 
\textcolor{black}{These results show that using LSTM for triggering and BERT for response generation outperforms other combinations in terms of average Precision@3 value of 85.58\%. 
We observe that using BERT for both phases leads to a very similar performance with an average Precision@3 value of 85.42\%, coming second among all the tested combinations.
The higher performance for using LSTM in the first phase can be attributed to the particularly good performance of LSTM for the triggering task.
That is, LSTM outperforms BERT in three performance metrics out of seven that we report in Table~2, with precision values for the feasible messages exceeding that of the BERT model by 2.55\% on average.}
We also find that rule-based approaches and their combinations have significantly lower performance, pointing to the benefits of employing machine learning algorithms to create this end-to-end pipeline.
}


\begin{table}[!ht]
\centering
\caption{\textcolor{black}{Summary performance values for the triggering and response generation model combinations to create an end-to-end pipeline.}}\label{tab:best_pipline}
\resizebox{\linewidth}{!}{
    \begin{tabular}{llrllllr}
    \toprule
    \textbf{\begin{tabular}[c]{@{}l@{}}Triggering\\ Model\end{tabular}} & \textbf{\begin{tabular}[c]{@{}l@{}}Response Generation\\ Model\end{tabular}} & \textbf{Precision@3} & \textbf{} & \textbf{} & \textbf{\begin{tabular}[c]{@{}l@{}}Triggering\\ Model\end{tabular}} & \textbf{\begin{tabular}[c]{@{}l@{}}Response Generation\\ Model\end{tabular}} & \textbf{Precision@3} \\
    \midrule
    \textbf{BERT} & \textbf{BERT} & 85.42 $\pm$ 0.82 &  &  & \cellcolor[HTML]{EFEFEF}\textbf{LSTM} & \cellcolor[HTML]{EFEFEF}\textbf{BERT} & \cellcolor[HTML]{EFEFEF} \textbf{85.58 $\pm$ 0.42} \\
    \textbf{BERT} & \textbf{LSTM} & 82.79 $\pm$ 0.81 &  &  & \textbf{LSTM} & \textbf{LSTM} & 83.28 $\pm$ 0.75 \\
    \textbf{BERT} & \textbf{Seq2seq} & 60.22 $\pm$ 4.80 &  &  & \textbf{LSTM} & \textbf{Seq2seq} & 61.56 $\pm$ 4.08 \\
    \textbf{BERT} & \textbf{XGBoost} & 80.89 $\pm$ 1.07 &  &  & \textbf{LSTM} & \textbf{XGBoost} & 81.92 $\pm$ 0.57 \\
    \textbf{BERT} & \textbf{SVM} & 81.11 $\pm$ 1.23 &  &  & \textbf{LSTM} & \textbf{SVM} & 82.03 $\pm$ 0.66 \\
    \textbf{BERT} & \textbf{Weighted-TFIDF} & 49.73 $\pm$ 4.03 &  &  & \textbf{LSTM} & \textbf{Weighted-TFIDF} & 51.63 $\pm$ 3.27 \\
    \textbf{BERT} & \textbf{TFIDF} & 50.33 $\pm$ 3.22 &  &  & \textbf{LSTM} & \textbf{TFIDF} & 52.13 $\pm$ 2.90 \\
    \textbf{BERT} & \textbf{Frequency} & 46.32 $\pm$ 1.82 &  &  & \textbf{LSTM} & \textbf{Frequency} & 48.18 $\pm$ 1.24 \\
    \textbf{XGBoost} & \textbf{BERT} & 83.61 $\pm$ 0.41 &  &  & \textbf{SVM} & \textbf{BERT} & 83.00 $\pm$ 0.39 \\
    \textbf{XGBoost} & \textbf{LSTM} & 81.37 $\pm$ 0.66 &  &  & \textbf{SVM} & \textbf{LSTM} & 80.68 $\pm$ 0.62 \\
    \textbf{XGBoost} & \textbf{Seq2seq} & 59.32 $\pm$ 4.27 &  &  & \textbf{SVM} & \textbf{Seq2seq} & 58.59 $\pm$ 4.19 \\
    \textbf{XGBoost} & \textbf{XGBoost} & 79.60 $\pm$ 0.52 &  &  & \textbf{SVM} & \textbf{XGBoost} & 78.83 $\pm$ 0.48 \\
    \textbf{XGBoost} & \textbf{SVM} & 79.81 $\pm$ 0.57 & \textbf{} & \textbf{} & \textbf{SVM} & \textbf{SVM} & 78.98 $\pm$ 0.49 \\
    \textbf{XGBoost} & \textbf{Weighted-TFIDF} & 48.85 $\pm$ 3.42 &  &  & \textbf{SVM} & \textbf{Weighted-TFIDF} & 47.79 $\pm$ 3.30 \\
    \textbf{XGBoost} & \textbf{TFIDF} & 49.45 $\pm$ 2.48 &  & \textbf{} & \textbf{SVM} & \textbf{TFIDF} & 48.41 $\pm$ 2.41 \\
    \textbf{XGBoost} & \textbf{Frequency} & 45.61 $\pm$ 0.63 &  &  & \textbf{SVM} & \textbf{Frequency} & 44.51 $\pm$ 0.46 \\
    \textbf{Weighted-TFIDF} & \textbf{BERT} & 73.31 $\pm$ 0.96 &  &  & \textbf{TFIDF} & \textbf{BERT} & 73.27 $\pm$ 0.66 \\
    \textbf{Weighted-TFIDF} & \textbf{LSTM} & 71.45 $\pm$ 0.91 &  &  & \textbf{TFIDF} & \textbf{LSTM} & 71.28 $\pm$ 0.72 \\
    \textbf{Weighted-TFIDF} & \textbf{Seq2seq} & 52.34 $\pm$ 3.12 &  &  & \textbf{TFIDF} & \textbf{Seq2seq} & 52.09 $\pm$ 3.31 \\
    \textbf{Weighted-TFIDF} & \textbf{XGBoost} & 69.96 $\pm$ 0.87 &  &  & \textbf{TFIDF} & \textbf{XGBoost} & 69.75 $\pm$ 0.68 \\
    \textbf{Weighted-TFIDF} & \textbf{SVM} & 70.14 $\pm$ 0.78 &  &  & \textbf{TFIDF} & \textbf{SVM} & 69.90 $\pm$ 0.56 \\
    \textbf{Weighted-TFIDF} & \textbf{Weighted-TFIDF} & 46.46 $\pm$ 2.97 &  &  & \textbf{TFIDF} & \textbf{Weighted-TFIDF} & 45.38 $\pm$ 2.90 \\
    \textbf{Weighted-TFIDF} & \textbf{TFIDF} & 44.44 $\pm$ 1.20 &  &  & \textbf{TFIDF} & \textbf{TFIDF} & 44.35 $\pm$ 1.28 \\
    \textbf{Weighted-TFIDF} & \textbf{Frequency} & 42.24 $\pm$ 0.74 &  &  & \textbf{TFIDF} & \textbf{Frequency} & 40.93 $\pm$ 0.70 \\
    \textbf{Frequency} & \textbf{BERT} & 61.87 $\pm$ 0.57 &  &  & \textbf{Frequency} & \textbf{SVM} & 58.17 $\pm$ 0.67 \\
    \textbf{Frequency} & \textbf{LSTM} & 59.92 $\pm$ 0.61 &  &  & \textbf{Frequency} & \textbf{Weighted-TFIDF} & 33.48 $\pm$ 2.82 \\
    \textbf{Frequency} & \textbf{Seq2seq} & 42.12 $\pm$ 3.18 &  &  & \textbf{Frequency} & \textbf{TFIDF} & 34.00 $\pm$ 1.97 \\
    \textbf{Frequency} & \textbf{XGBoost} & 57.98 $\pm$ 0.69 &  &  & \textbf{Frequency} & \textbf{Frequency} & 32.20 $\pm$ 0.67\\
    \bottomrule
    \end{tabular}
}
\end{table}

\end{document}